\documentclass[twocolumn]{IEEEtran}

\usepackage[numbers]{natbib}
\usepackage{amssymb}
\usepackage{amsmath}
\usepackage{hyperref}
\hypersetup{colorlinks=true, linkcolor=blue, anchorcolor=blue, citecolor=blue}
\usepackage{amssymb} 
\usepackage{amsmath}  
\usepackage{caption} 
\usepackage{graphicx}
\usepackage{float} 
\usepackage{subcaption}
\usepackage[export]{adjustbox} 
\usepackage[ruled,linesnumbered,lined]{algorithm2e}
\usepackage{longtable}
\usepackage{multirow}
\usepackage{tabularx}
\usepackage{booktabs}
\usepackage[table]{xcolor}

\def\tsc#1{\csdef{#1}{\textsc{\lowercase{#1}}\xspace}}
\tsc{WGM}
\tsc{QE}
\tsc{EP}
\tsc{PMS}
\tsc{BEC}
\tsc{DE}

\begin{document}
\let\printorcid\relax
\let\WriteBookmarks\relax
\def\floatpagepagefraction{1}
\def\textpagefraction{.001}

\title{WIFE-Fusion:Wavelet-aware Intra-inter Frequency Enhancement for Multi-model Image Fusion}                      


\author{
  Tianpei~Zhang,
  Jufeng~Zhao,
  Yiming~Zhu,
  Guangmang~Cui

  \thanks{
    This work was supported by
      the National Natural Science Foundation of China (xxxxx),

    \emph{(Corresponding author: xxxxxx)}
    }

}
\maketitle










\begin{abstract}
Multimodal image fusion effectively aggregates information from diverse modalities, with fused images playing a crucial role in vision systems. However, existing methods often neglect frequency-domain feature exploration and interactive relationships. In this paper, we propose wavelet-aware Intra-inter Frequency Enhancement Fusion (WIFE-Fusion), a multimodal image fusion framework based on frequency-domain components interactions. Its core innovations include: Intra-Frequency Self-Attention (IFSA) that leverages inherent cross-modal correlations and complementarity through interactive self-attention mechanisms to extract enriched frequency-domain features, and Inter-Frequency Interaction (IFI) that enhances enriched features and filters latent features via combinatorial interactions between heterogeneous frequency-domain components across modalities. These processes achieve precise source feature extraction and unified modeling of feature extraction-aggregation. Extensive experiments on five datasets across three multimodal fusion tasks demonstrate WIFE-Fusion's superiority over current specialized and unified fusion methods. Our code is available at https://github.com/Lmmh058/WIFE-Fusion.
\end{abstract}



\begin{IEEEkeywords}
Multimodal Image Fusion, Frequency-domain Interaction, Wavelet Transform.
\end{IEEEkeywords}

\section{Introduction} \label{sec:introduction}
Each imaging sensor can capture information from different dimensions due to their unique imaging mechanisms \cite{ma2019infrared}. 
Visible light (VIS) and near-infrared (NIR) sensors capture images by detecting light reflected from objects, while infrared (IR) sensors generate thermal images by measuring self-emitted thermal radiation from targets.
In the medical image analysis, Positron Emission Tomography (PET) generates images by monitoring the distribution of radioactive tracers, while Computed Tomography (CT) measures X-ray attenuation differences across tissues to create anatomical images. Magnetic Resonance Imaging (MRI) employs magnetic fields and radio waves to produce high-resolution cross-sectional images of internal bodily structures. 
Notably, images captured under different sensors or multiple shooting settings typically contain complementary information \cite{wang2023mct,li2020multi}.

For instance, VIS sensors capture rich textural details but are highly susceptible to illumination variations, adverse weather conditions, and occlusions. In contrast, IR sensors provide stable thermal imaging across diverse environmental conditions, yet lack textural specificity. NIR sensors complement visible imagery under low-light or obscured scenarios by enhancing detail representation.
Moreover, PET provides abundant metabolic activity information, CT demonstrates superior imaging capabilities for high-density tissues, while MRI delivers high-resolution anatomical details.
Therefore, there is an urgent need to fuse images from different imaging mechanisms to significantly improve overall image quality and information content.
This is of great significance for practical application fields such as intelligent monitoring \cite{zhu2024towards}, autonomous driving \cite{feng2020deep}, and medical diagnosis \cite{goyal2022multi}. 
Consequently, this paper particularly focuses on the key technology research of infrared and visible image fusion (IVIF), RGB and near-infrared image fusion (RGBNIR) and medical(MIF) image fusion.

Over the past few decades, numerous methods for multi-modal image fusion have emerged, which can be divided into traditional methods and deep learning methods. Traditional methods mainly rely on handcrafted features and fusion rules derived from domain expertise. This includes multi-scale transformation (MST) \cite{selvaraj2020infrared}, sparse representation \cite{yang2020infrared}, subspace models \cite{lu2014novel} and saliency-based methods \cite{ma2017infrared}. Despite these methods can achieve satisfactory results in specific scenarios, they still struggle in dealing with complex scenes.

With the flourish of deep learning technology, an increasing number of deep learning methods utilizing autoencoders (AE) \cite{li2021rfn,xie2025scdfuse,mei2024gtmfuse}, convolutional neural networks (CNNs) \cite{xu2020u2fusion, zhang2025daaf}, and generative adversarial networks (GANs) \cite{ma2019fusiongan} have achieved enhanced performance and generalization capabilities. However, due to the limited receptive fields of convolutional operations, many deep learning approaches incorporate multi-scale pyramid architectures \cite{wang2021unfusion}, attention mechanisms \cite{tang2023datfuse}, or other techniques to expand the effective receptive field. Additionally, Vision Transformer (ViT)-based methods \cite{wang2022swinfuse} model relationships among all pixels through self-attention mechanisms to capture long-range dependence features. The ViT overcome the local receptive field of convolution operations, but still facing bottleneck limitations.

Specifically, existing ViT methods \cite{hu2025datransnet,wang2024aitfuse} have achieved promising results in image fusion due to their excellent long-range dependency modeling relationships.
However, most methods, such as DATFuse \cite{tang2023datfuse} and SwinFusion \cite{ma2022swinfusion}, only consider spatial domain features and fail to fully consider frequency domain features, which limits the improvement of fusion performance in multiple aspects. 
Firstly, existing methods typically concentrate on the complex multi-level process of applying ViT in the spatial domain, including feature extraction, interaction, and reconstruction. However, feature extraction in a single spatial domain can easily lead to significant information loss at small edges and other locations. 
Secondly, most current methods lack the enhancement and interaction of frequency domain features. For instances, the Spatial Frequency Information Integration Network \cite{zhou2024general} enhances the learning ability of the model by combining local spatial information and global frequency information, and achieves excellent performance in panoramic sharpening and deep super-resolution tasks. 
However, it fails to simultaneously interact between low-frequency structural brightness information and high-frequency detail information across different modalities. This shortcoming can lead to the network overlooking the deep relationships between modality-specific structures, details, and their interactions, thereby preventing the network from obtaining a comprehensive representation of image features across different modalities.

\begin{figure}
    \centering
    \includegraphics[width=\columnwidth]{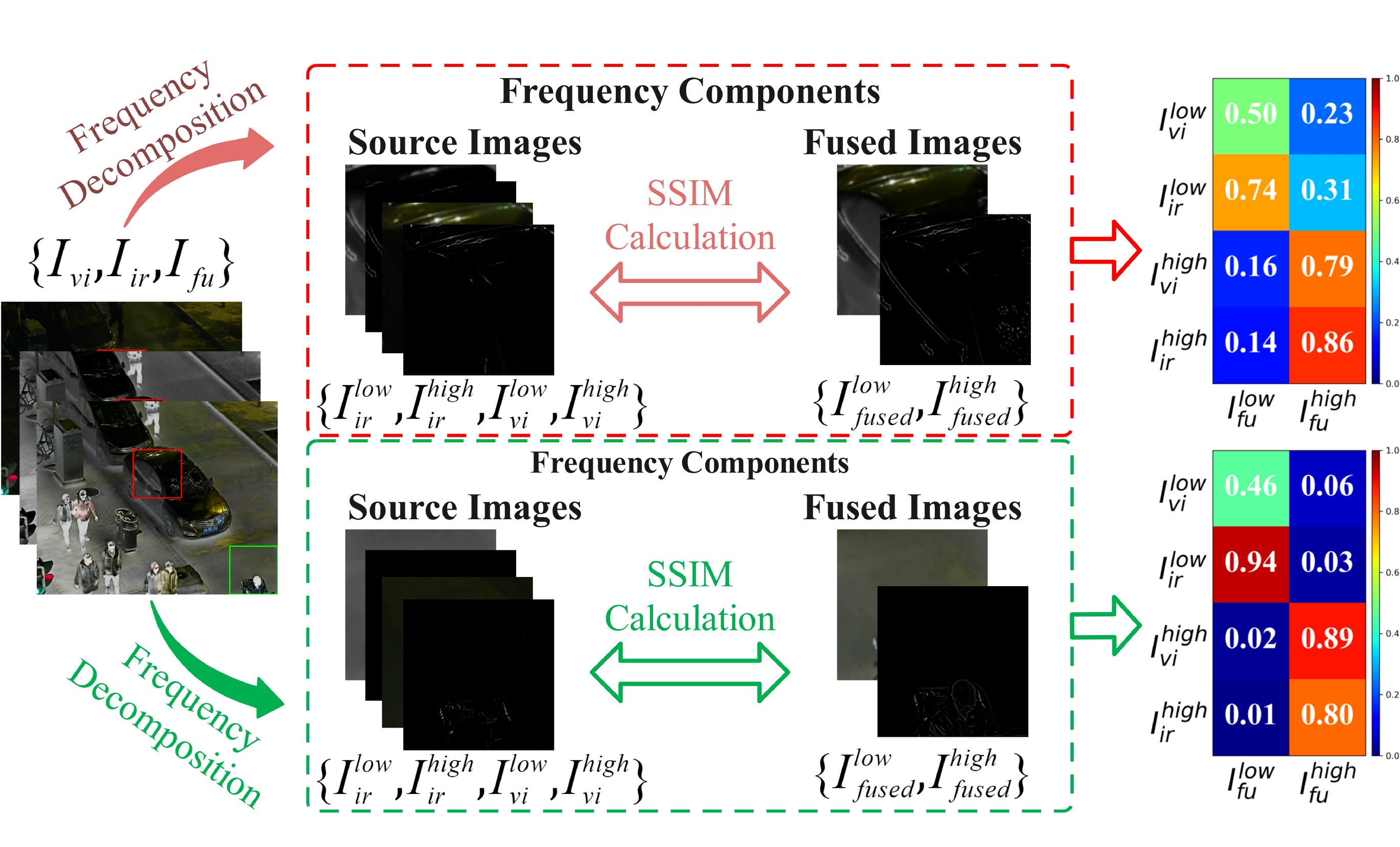}
    \caption{SSIM computation results between the frequency components of the source and fused images after applying frequency decomposition to the sample images.}
    \label{fig:fig1}
\end{figure}

We argue the key to multi-modal fusion lies in fully extracting and interacting through frequency domain features. Specifically, Frequency domain features can reveal frequency multi-modal image details that are difficult to extract using current deep learning methods.
As illustrated in Fig. \ref{fig:fig1}, we performed frequency decomposition on selected \textcolor{red}{red} boxed and \textcolor{green}{green} boxed from visible, infrared, and fused images, followed by SSIM similarity calculations between the four frequency components of source images and the two frequency components of fused images. In the \textcolor{green}{green} boxed region, the fused image's frequency components demonstrate high correlation with same-frequency components of source images but low correlation with cross-frequency components, indicating predominant information transfer from same-frequency counterparts. In contrast, within the \textcolor{red}{red} boxed region, the frequency components of the fused image not only exhibit strong correlations with the same-frequency components of the source images but also demonstrate associations with cross-frequency components. 

This insightful discovery confirms our viewpoint of fully extracting and interacting through frequency domain features, which indicates the existence of inter-frequency information transfer in specific regions, where the required information in the current frequency band may retain a residual presence in cross-frequency components. Therefore, the necessity of implementing high-low frequency feature interactions in multi-modal image fusion to supplement frequency-enriched features and re-exploit latent information through cross-frequency collaboration. The future of multi-modal fusion should concentrate on frequency domain features, especially the interaction between high-frequency and low-frequency features, to further enhance the performance of multi-modal fusion and image fusion.

To address the aforementioned issues, we propose a multi-modal image fusion framework, \textbf{W}avelet-aware \textbf{I}ntra-inter \textbf{F}requency \textbf{E}nhancement Fusion (WIFE-Fusion). Specifically, the method first decomposes multi-modal images into high- and low-frequency components using wavelet transforms. We then proposed two core models: \textbf{I}ntra-\textbf{F}requency \textbf{S}elf-\textbf{A}ttention (IFSA) and \textbf{I}nter-\textbf{F}requency \textbf{I}nteraction (IFI). Our IFSA facilitates comprehensive interactions by aligning the queries and values derived from frequency features with keys from cross-modal same-frequency components. This symmetric interactive attention mechanism effectively detects complementarity and correlations within identical frequency components across modalities. Moreover, the proposed IFI establishes cross-frequency information pathways through focused attention on critical intra-frequency features and combinatorial interactions with heterogeneous frequency components, achieving enriched feature supplementation and residual feature filtering in the frequency domain. WIFE-Fusion enhances the accurate capture of global spatial relationships and holistic structures, thereby generating fused images with reinforced textural details.

The core contributions of this research can be described as follows:

\begin{enumerate}
    \item We have developed a novel multi-modal image fusion framework named WIFE-Fusion, which operates through wavelet-domain frequency feature extraction while emphasizing cross-modal and cross-frequency interaction relationships to generate fused images with rich information and superior visual quality.
    \item In WIFE-Fusion, we designed Intra-Frequency Self-Attention (IFSA) and Inter-Frequency Interaction (IFI) to explore cross-modal correlations and complementarity within identical frequency components and establish information pathways across heterogeneous-frequency components between modalities, respectively.
    \item Extensive experimental results across five datasets spanning three multi-modal image fusion tasks demonstrate that our WIFE-Fusion outperforms state-of-the-art specialized and unified fusion methods.
\end{enumerate}

\section{Related Work} \label{sec:related}

In this section, we firstly reviewed some representative studies in multi-modal image fusion. Then, representative works combining wavelet transform and deep learning were reviewed and discussed.

\subsection{Multi-modal Image Fusion}
\subsubsection{Traditional Methods}

Traditional methods primarily rely on handcrafted features and fusion rules derived from domain expertise. In the IVIF, these methods compass multi-scale transformations \cite{burt1987laplacian,zhan2017infrared,li2016infrared,selvaraj2020infrared}, using curvelet transform \cite{nencini2007remote,ali2010curvelet}, nonsubsampled contourlet transform \cite{xiang2015fusion,zhu2019phase}, wavelet transform \cite{bhavana2015multi, hill2016perceptual}, and cosine transform \cite{wang2020fast}. These techniques are employed to extract salient features from infrared and visible images. Additionally, sparse representation methods \cite{wang2022medical, wang2014fusion,lu2014infrared,yang2020infrared} utilize latent low-rank representations and dictionary learning to enhance feature extraction. The subspace method \cite{li2023infrared,lu2014novel,zhang2014multi} project high-dimensional images into low dimensional subspaces. The saliency method \cite{han2013fast, ma2017infrared, cui2015detail,liu2022infrared} detect significant object regions to optimize feature representation and guide feature combination through weight maps. In the MIF, the traditional to obtaining multi-modal fusion methods include Laplace transform \cite{du2017anatomical}, curve wave transform \cite{ali2010curvelet}, wavelet transform \cite{bhavana2015multi}, contour wave transform \cite{zhu2019phase}, sparse representation \cite{wang2022medical}, subspace method \cite{kim2016joint}, saliency method \cite{han2013fast}, These methods have been widely adopted for their ability to capture essential information from different modalities.

Although these traditional methods can achieve satisfactory fusion results in specific scenarios, they have notable limitations. They often require complex transformations and struggle to adapt to different scenes. Furthermore, they frequently apply uniform decomposition approaches to different image modalities without considering their intrinsic characteristics. This can result in inaccurate and inadequate feature extraction and suboptimal fusion outcomes.

\subsubsection{Deep-Learning Methods}

With the development of large-scale image datasets, deep learning methods have achieved satisfactory results in multi-modal image fusion tasks.

In the application of IVIF and RGBNIR fusion, several notable contributions have been made. For instance, Li \textit{et al.} proposed RFN-Nest \cite{li2021rfn}, which introduces a residual fusion network instead of manually designed fusion strategy. Liu \textit{et al.} \cite{liu2023sgfusion} proposed SGFusion, which utilizes saliency masks to emphasize saliency regions. Xu \textit{et al.} \cite{xu2020u2fusion} developed U2Fusion framework, where different fusion tasks are designed to complement each other, enhancing overall fusion quality. Ma \textit{et al.} \cite{ma2019fusiongan} pioneered the integration of GANs into image fusion, with the generator crafting fused images that incorporate both visible light gradients and infrared intensity data. 
Furthermore, based on the Vision Transformer (ViT) architecture \cite{dosovitskiy2020image}, Tang et al. \cite{tang2023datfuse} proposed DATFuse, which incorporates a dual-attention residual module to extract salient features and enhance the fusion performance for the IVIF task.

In the application of MIF, Liu \textit{et al.} \cite{liu2017medical} proposed a medical image fusion framework with twin CNNs. The SDNet \cite{zhang2021sdnet} is a sufficiently real-time and squeeze-and-decomposition network has been developed. The EMFuion\cite{xu2021emfusion} is an unsupervised enhanced fusion network that utilizes surface layer and deep layer constraints to improve information preservation. Moreover, Ma \textit{et al.} proposed DDcGAN \cite{ma2020ddcgan}, an adversarial network was established between a generator and two discriminators. Xie \textit{et al.} proposed FusionMamba \cite{xie2024fusionmamba}, the first multimodal image fusion framework based on the Mamba architecture \cite{gu2023mamba}.

Our WIFE-Fusion has several key improvements compared to other methods:

\begin{enumerate}
    \item \textbf{Frequency-domain structure-detail decoupling:} Existing methods primarily focus on spatial-domain feature extraction and enhancement \cite{tang2023datfuse} while neglecting the insufficient feature extraction caused by the coupling of structural and detailed information in the spatial domain. Our WIFE-Fusion achieves structure-detail decoupling through wavelet transforms and enables comprehensive extraction of structural and detailed information via interaction mechanisms.

    \item \textbf{Frequency-domain cross-modal complementary enhancement:} Existing methods achieve cross-modal complementary enhancement through spatial-domain features \cite{ma2022swinfusion}, but the information coupling in spatial domain may limit the capacity for deep-level feature interactions. The Wavelet-aware Intra-inter Frequency Enhancement module (WIFE) effectively enhances deep interactions and feature fusion via intra- and inter-frequency component interactions.
\end{enumerate}

\subsection{Wavelet Transform for Image Fusion}

The integration of deep learning and wavelet transform has made significant progress in image fusion applications. Wavelet transform can simultaneously analyze signals in both time and frequency domains, making it a powerful tool in multi-image fusion. Through wavelet transform, images can be decomposed into multiple sub-bands at different resolutions, and the information from these sub-bands can be fused as required.  
For instance, wavelet-based self-supervised learning \cite{liu2022wavelet} decompose the image into sub-bands of different frequencies through wavelet transform, and then use a self-supervised learning framework to optimize the fusion rules.
MFIF-DWT-CNN \cite{avci2024mfif} performs Discrete Wavelet Transform (DWT) decomposition on the input image and subsequently inputs the decomposed coefficients into CNNs for feature extraction and fusion. By combining the multi-scale analysis capability of wavelet transform with the deep learning capabilities of CNNs, the clarity and detail preservation of the fused image have been significantly enhanced.

However, although wavelet transforms combined with convolutional neural networks have achieved certain progress in image fusion quality, existing methods still lack sufficient attention to the full utilization of frequency-domain decoupled information and the establishment of interaction pathways between frequency components. Compared with existing methods, our contributions are highlighted as follows:

\begin{enumerate}
    \item \textbf{Frequency-Aware Transformer for Robust Feature Learning}: Most frequency-domain fusion frameworks \cite{wu2024dcfnet} use shallow networks or convolution-based modules, which struggle to capture long-range dependencies. In contrast, our WIFE incorporate Transformer architectures to model both intra-frequency long-range dependencies and inter-frequency complementarity.

    \item \textbf{Cross-modal cross-frequency feature interaction:} Current research often overlooks cross-frequency interactions between high and low frequencies, potentially hindering information flow across frequency components. In contrast, our Inter-Frequency Interaction module establishes information pathways between frequency components through frequency-aware attention and combinatorial interactions, leveraging complementary characteristics between components to acquire effective residual information.
\end{enumerate}

\begin{figure*}[!t]
    \centering
    \includegraphics[width=1\textwidth]{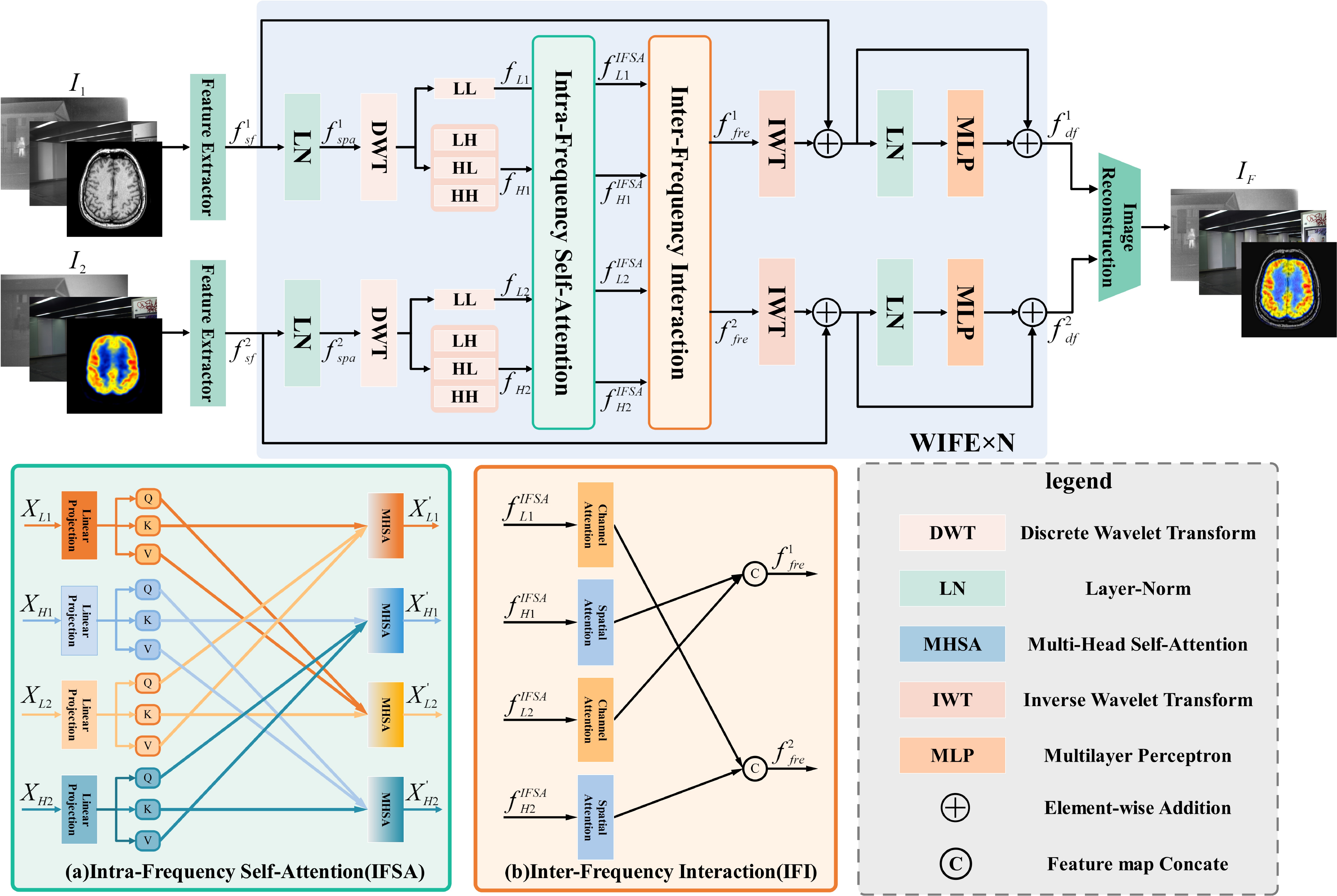}
    \caption{Overall architecture of the proposed WIFE-Fusion framework. The network consists of \(N\) sequential WIFE modules, with the specific structures of the Intra-Frequency Self-Attention and Inter-Frequency Interaction illustrated in (a) and (b), respectively.}
    \label{fig:network}
\end{figure*}

\section{Method} \label{sec:method}

In Sec. \ref{sec:3a}, we first introduce the overall framework of the Wavelet Interactive Feature Exchange Fusion (WIFE-Fusion). In Sec. \ref{sec:3b}, we present detailed explanations of the Wavelet-aware Intra-inter Frequency Enhancement module (WIFE) and its two core components: Intra-frequency Self-Attention (IFSA) and Inter-frequency Interaction (IFI). Finally, in Sec. \ref{sec:3c}, we present an overview of the loss function design.

\subsection{Overall FrameWork}
\label{sec:3a}
The overall network architecture of WIFE-Fusion is illustrated in Fig. \ref{fig:network}. Source images from two different modalities, each with dimensions \(I_{1},I_{2} \in \mathbb{R}^{H \times W \times 1}\), are input into the network through two separate branches. It should be noted that when processing RGB input images, we first convert them to the YCbCr color space and extract the Y component as the network input. The fusion result from the network output is then combined with the original Cb and Cr components before converting back to the RGB color space, producing the final color fused image.

Firstly, shallow features are extracted using a three-layer lightweight CNNs , enabling the acquisition of more meaningful feature representations:
\begin{equation}
\label{eq:ShallowFeatureEx}
f_{sf}^{1} = FE(I_{1}),f_{sf}^{2} = FE(I_{2})
\end{equation}

\noindent where \( f_{sf}^{1}, f_{sf}^{2} \in \mathbb{R}^{B \times C \times H \times W} \) represent the shallow feature maps of the two modalities obtained through feature extractor \( \text{FE}(\cdot) \). These feature maps are then fed into \( N \) cascaded Wavelet-aware Intra-inter Frequency Enhancement (WIFE) modules, which perform frequency feature extraction and intra-inter frequency complementary feature enhancement:
\begin{equation}
\label{eq:DeepFeatureEx}
f_{df}^{1} = WIFE^{N}(f_{sf}^{1}),f_{df}^{2} = WIFE^{N}(f_{sf}^{2})
\end{equation}

\noindent where \( WIFE^{N}(\cdot) \) represents the \( N \) cascaded WIFE modules. \( f_{df}^{1}, f_{df}^{2} \in \mathbb{R}^{B \times C \times H \times W}\) denote the deep features for the two modalities. Subsequently, \( f_{df}^{1}\) and \(f_{df}^{2} \) are input into a CNN-based feature fusion and image reconstruction module to obtain the final fused result \( I_F \), which can be calculated as follows:
\begin{equation}
\label{eq:Fusion&Reconstruction}
I_{F} = Fu\&Re(f_{df}^{1},f_{df}^{2})
\end{equation}

\noindent where \( I_{F} \in \mathbb{R}^{H \times W \times 1}\) represents the multi-modal fused image, and \( Fu\&Re(\cdot) \) consists of three sequential convolution layers (with a kernel size of \( 3 \times 3 \)) followed by Leaky-ReLU activation functions.

\subsection{Wavelet-aware Intra-inter Frequency Enhancement Module}
\label{sec:3b}
Firstly, despite the features \( f_{sf}^{1}, f_{sf}^{2} \in \mathbb{R}^{B\times C\times H\times W}\) extracted by \(FE(\cdot) \) contain rich local features and spatial domain information. However, they still lack of the modeling of long-range dependencies within the feature maps and the thorough exploration of frequency-domain feature information. Secondly, during the previous process, the features from different modalities are processed independently, and the feature correlations and complementarities between modalities have not been fully explored. To address the aforementioned two issues, we propose WIFE, whose structure is illustrated in Fig. \ref{fig:network}. WIFE consists of two core components: intra-frequency self-attention (IFSA) and inter-frequency interaction (IFI), which work together to achieve the desired functionality. The workflow of WIFE can be defined as:
\begin{equation}
\label{eq:2DDWT}
\begin{aligned}
&f_{spa}^{1} = LN(f_{sf}^{1}),f_{spa}^{2} = LN(f_{sf}^{2})\\
&f_{fre}^{1} = IFI(IFSA(DWT(f_{spa}^{1}),DWT(f_{spa}^{2}))) \\
&f_{fre}^{2} = IFI(IFSA(DWT(f_{spa}^{2}),DWT(f_{spa}^{1}))) \\
&f^{1'} = IWT(f_{fre}^{1}) + f_{sf}^{1}, f^{2'} = IWT(f_{fre}^{2}) + f_{sf}^{2} \\
&f_{out}^{1} = MLP(LN(f^{1'})) + f^{1'} \\
&f_{out}^{2} = MLP(LN(f^{2'})) + f^{2'}
\end{aligned}
\end{equation}

\noindent where \( f_{sf}^{1}, f_{sf}^{2} \in \mathbb{R}^{B \times C \times H \times W} \) represent the input features of WIFE, \(f_{spa}^{1}, f_{spa}^{2} \in \mathbb{R}^{B \times C \times H \times W}\) represent the spatial-domain features of two modalities after layer normalization (\(LN(\cdot)\)), while \( DWT(\cdot), IWT(\cdot), IFSA(\cdot) \) and \( IFI(\cdot) \) denote discrete wavelet transform, inverse wavelet transform, \textbf{I}ntra-\textbf{F}requency \textbf{S}elf-\textbf{A}ttention(IFSA) and \textbf{I}nter-\textbf{F}requency \textbf{I}nteraction(IFI), respectively. \(f_{fre}^{1}, f_{fre}^{2} \in \mathbb{R}^{4B \times C \times H' \times W'}\) (\(H' = \frac{H}{2}, W' = \frac{W}{2}\)) represent the wavelet frequency-domain features of two modalities after deep interactive processing. \( f_{out}^{1}, f_{out}^{2} \in \mathbb{R}^{B \times C \times H \times W} \) represent the output features of a single WIFE.

\textbf{Intra-frequency Self-Attention (IFSA)}: 
To fully exploit cross-modal feature correlations and complementarity to acquire essential structural and detailed information, we propose Intra-Frequency Self-Attention (IFSA) to establish cross-modal frequency component interactions, which effectively explores latent structural and detailed relationships across modalities. For the implementation of IFSA, we first transform spatial-domain features into frequency-domain representations via wavelet transforms, to decompose them into distinct frequency bands. These frequency bands components are composed of structural/brightness information (low-frequency components) and detail information (high-frequency components), which can be expressed as:
\begin{equation}
\label{eq:2DDWT}
\begin{aligned}
LL^{1},LH^{1},HL^{1},HH^{1} = DWT(f_{spa}^{1}) \\
LL^{2},LH^{2},HL^{2},HH^{2} = DWT(f_{spa}^{2})
\end{aligned}
\end{equation}

\noindent where \(LL^{1}, LL^{2} \in \mathbb{R}^{B \times C \times H' \times W'}\) (\(H' = \frac{H}{2}, W' = \frac{W}{2}\)) denote their low-frequency components, and \(LH^{1}, HL^{1}, \allowbreak HH^{1}, LH^{2}, HL^{2}, HH^{2} \in \mathbb{R}^{B \times C \times H' \times W'}\) correspond to the high-frequency components. We then concatenate the three high-frequency subbands along the batch dimension, obtaining \(f_{H1} = concat(LH^{1}, HL^{1}, HH^{1})\) and \(f_{H2} = concat(LH^{2}, HL^{2}, HH^{2})\) for subsequent processing. The low-frequency features are preserved as \(f_{L1} = LL^{1}\) and \(f_{L2} = LL^{2}\).

After computing \(f_{H1}\), \(f_{H2} \in \mathbb{R}^{3B \times C \times H' \times W'}\) and \(f_{L1}, \allowbreak f_{L2} \in \mathbb{R}^{B \times C \times H' \times W'}\), we perform self-attention calculations between the queries \(Q\) and values \(V\) corresponding to the frequency-domain components and the keys \(K\) from cross-modal same-frequency components, which can be formally defined as:

\begin{equation}
\label{eq:IFSA}
\begin{aligned}
\{Q_{L1},K_{L1},V_{L1}\}  &= \{X_{L1}W^{Q}_{L1},X_{L1}W^{K}_{L1},X_{L1}W^{V}_{L1}\} \\
\{Q_{L2},K_{L2},V_{L2}\}  &= \{X_{L2}W^{Q}_{L2},X_{L2}W^{K}_{L2},X_{L2}W^{V}_{L2}\} \\
X'_{L1} &= MHSA(Q_{L2},K_{L1},V_{L2}) \\
X'_{L2} &= MHSA(Q_{L1},K_{L2},V_{L1}) 
\end{aligned}
\end{equation}

\noindent where \( X_{L1}, X_{L2} \in \mathbb{R}^{W^2 \times C} \) represent the individual window vectors obtained from the window partition or shifted window partition operations applied to \( f_{L1} \) and \( f_{L2} \), respectively. \( MHSA(\cdot) \) refers to the Multi-Head Self-Attention operation \cite{vaswani2017attention}, and \( X'_{L1}, X'_{L2} \in \mathbb{R}^{W^2 \times C} \) are the output vectors of the individual windows obtained via the IFSA. These vectors are then reshaped into the output features of the IFSA, denoted as \( f_{L1}^{IFSA}, f^{IFSA}_{L2} \in \mathbb{R}^{B \times C \times H' \times W'}\). Similarly, through the same process as in Eq. \ref{eq:IFSA}, we can obtain the high-frequency features of the IFSA output, denoted as \( f^{IFSA}_{H1}, f^{IFSA}_{H2} \in \mathbb{R}^{3B \times C \times H' \times W'}\).

\textbf{Inter-frequency Interaction (IFI)}:
To establish information pathways between cross-modal heterogeneous frequency features and address the inadequate extraction of enriched features and residual feature retention in frequency components, we propose Inter-Frequency Interaction (IFI). Specifically, we employ channel-spatial attention mechanisms \cite{woo2018cbam} to highlight critical information within specific frequency bands, while performing cross-frequency interactions between cross-modal heterogeneous frequencies to achieve deep complementary relationships across modalities and frequencies, simultaneously enabling feature fusion between modalities. The IFI process can be formally defined as:
\begin{equation}
\label{eq:IFI}
\begin{aligned}
&f_{fre}^{1} = concat(CA(f^{IFSA}_{L1}),SA(f^{IFSA}_{H2})))  \\
&f_{fre}^{2} = concat(CA(f^{IFSA}_{L2}),SA(f^{IFSA}_{H1}))) \\
\end{aligned}
\end{equation}

\noindent where \(f_{fre}^{1}, f_{fre}^{2} \in \mathbb{R}^{4B \times C \times H' \times W'}\) represent the frequency-domain interaction features of two modalities processed by IFI, \(concat(\cdot)\) denotes the concatenation operation along the batch dimension, and \(CA(\cdot)\) and \(SA(\cdot)\) correspond to channel attention and spatial attention mechanisms respectively, which are defined as follows:
\begin{equation}
\label{eq:Attention Mechanism}
\begin{aligned}
&CA(x) = \sigma(MLP(Avg(x)) + MLP(Max(x))) \otimes x\\
&SA(x) = \sigma(Conv_{7}^{2,1}(Con(Avg(x),Max(x)))) \otimes x\\
\end{aligned}
\end{equation}

\noindent where \( Max(\cdot) \) and \( Avg(\cdot) \) represent the max pooling and average pooling operations, respectively. \(MLP(\cdot)\) stands for multi-layer perceptron, \( \sigma \) denotes the Sigmoid activation function, and \( Con(\cdot) \) refers to the concatenation operation along the channel dimension. \( Conv_{7}^{2,1} \) is a convolution operation with a kernel size of 7, and input and output channels of 2 and 1, respectively.

\begin{table*}[!t]
  \setlength{\abovecaptionskip}{0cm}  
  \renewcommand\arraystretch{1.2}
  \footnotesize
  \centering
  \vspace{-1\baselineskip}
\caption{CONFIGURATIONS FOR ALL COMPARATIVE EXPERIMENTS}
\vspace{-1\baselineskip}
\label{tab:parameters}
\tabcolsep=0.3cm
\renewcommand\arraystretch{1.2}
\begin{center}
\begin{tabular}{ll}
\toprule[1pt]
Methods(\textit{Source \& Year \& Type})         & Key parameters configurations   \\ \hline
\multicolumn{2}{l}{\textit{Infrared and Visible Image Fusion Methods}}  \\ \hline
GANMcC \cite{ma2020ganmcc} \textit{TIM'2020, GAN-based} & Patch Size: 120 $\times$ 120, $b = 32, p = \frac{1}{2}, M = 10, \gamma = 100, \beta_{1} = 1, \beta_{2} = 5, \beta_{3} = 4, \beta_{4} = 0.3$ \\ 
DATFuse \cite{tang2023datfuse} \textit{TCSVT'2023, Transformer-based} & Patch Size: 120 $\times$ 120, $C = 16, R = 4, \alpha=1, \lambda = 100, \gamma = 10$ \\ 
W-Mamba \cite{zhang2025exploring} \textit{2025, Mamba-based} & Patch Size: 128 $\times$ 128, $\lambda_{1} = 10, \lambda_{2} = 1$\\\hline
\multicolumn{2}{l}{\textit{Medical Image Fusion Methods}}  \\ 
\hline
EMFusion \cite{xu2021emfusion} \textit{INFFUS'2021, AE-based}  & Patch Size: 24 $\times$ 24, $\alpha = 0.5, \tau = 0.85, \lambda = 3, \xi = 10, \beta = [0,10], \eta = [0.5,1]$\\
GeSeNet \cite{li2023gesenet} \textit{TNNLS'2023, CNN-based} & Patch Size: 24 $\times$ 24, $\alpha = 0.5, \beta = 0.3, \gamma = 10, \sigma = 0.5$ \\
INet \cite{he2025inet} \textit{INFFUS'2025, INN-based} & $\alpha = 3, \beta = 1, \gamma = 1, \sigma = 1, \omega = 0.01$ \\
\hline
\multicolumn{2}{l}{\textit{Unified Image Fusion Methods}}  \\ \hline 
IFCNN \cite{zhang2020ifcnn} \textit{INFFUS'2020, CNN-based} & Patch size: \{64 $\times$ 64, 32 $\times$ 32\}, $\omega_{1} = 1, \omega_{2} = 1$ \\ 
PMGI \cite{zhang2020rethinking} \textit{AAAI'2020, CNN-based} &    Patch size: \{60 $\times$ 60, 120 $\times$ 120\}\\
U2Fusion \cite{xu2020u2fusion} \textit{TPAMI'2020, CNN-based}   &   Patch size: 64 $\times$ 64, $\alpha = 20, \lambda = 8e4, c = [3e3,3.5e3,1e2]$                   \\
SDNet \cite{zhang2021sdnet} \textit{IJCV'2021, CNN-based}     &  Patch size: \{60 $\times$ 60, 120 $\times$ 120\}, $\alpha = [0.5,0.5,0.5,1,1], \beta = [10,80,50,3]$                    \\ 
SwinFusion \cite{ma2022swinfusion} \textit{JAS'2022, Transformer-based}    &  Patch size: 128 $\times$ 128, $\lambda_{1} = 10, \lambda_{2} = 20, \lambda_{3} = 20$                     \\ 
CDDFuse \cite{zhao2023cddfuse} \textit{CVPR'2023, AE-based}    &  Patch size: 128 $\times$ 128, $\alpha_{1} = 1, \alpha_{2} = 2, \alpha_{3} = 10, \alpha_{4} = 2$                     \\
Fusionmamba \cite{xie2024fusionmamba} \textit{2024, Mamba-based}     &    Patch size: \{256 $\times$ 256, 358 $\times$ 358\}, $\alpha_{1} = 100, \alpha_{2} = 10, \alpha_{3} = 1$                   \\
EMMA \cite{zhao2024emma} \textit{CVPR'2024, Transformer-based}     &   Patch size: 128 $\times$ 128, $\alpha_{1} = 1, \alpha_{2} = 0.1$               \\ 

\rowcolor[rgb]{0.9,0.9,0.9}$\star$ \textbf{WIFE-Fusion(Ours), Transformer-based} & Patch size: 128 $\times$ 128, $\alpha = 2, \beta = 10, \gamma = 1$ \\
\bottomrule[1pt]
\end{tabular}
\end{center}
\end{table*}

\subsection{Loss Function}
\label{sec:3c}
In the training of WIFE-Fusion, our objective is to ensure that the network retains sufficient complementary information from different modalities while preserving maximal details. Therefore, a multi-term loss function is designed, integrating intensity, texture, and SSIM-based constraints. The overall loss function is defined as:

\begin{equation}
\label{eq:Loss Function}
\mathcal{L} = \alpha \mathcal{L}_{Int} + \beta \mathcal{L}_{text} +\gamma \mathcal{L}_{SSIM}
\end{equation}

\noindent where \( \alpha \), \( \beta \), and \( \gamma \) are the weighted coefficients for the intensity loss function \( \mathcal{L}_{Int} \), texture loss function \( \mathcal{L}_{text} \), and SSIM loss function \( \mathcal{L}_{SSIM} \), respectively. 

The purpose of the intensity loss is to ensure the fused image retains sufficient intensity information from both inputs. This is achieved by calculating the L1 distance between the fusion result and each source image, as follows:
\begin{equation}
\label{eq:lossint}
  \mathcal{L}_{Int} = \frac{\alpha_{1}}{HW}||I_{F}-I_{1}||_{1}+\frac{\alpha_{2}}{HW}||I_{F}-I_{2}||_{1} 
\end{equation}

\noindent where $||\cdot||_{1}$ denotes the L1 norm, while $I_{F}$, $I_{1}$, and $I_{2}$ correspond to the fused image and the two input sources, respectively. $H$ and $W$ indicate the height and width of the images. The weights $\alpha_{1}$ and $\alpha_{2}$ are assigned to balance the pixel-wise errors between the fused result and each source image. To ensure equal contribution from both inputs in terms of intensity, we set $\alpha_{1} = \alpha_{2} = 1$.

The texture loss function is designed to guide the network to incorporate richer texture details from the source images. Thus, the texture loss function can be defined as follows:
\begin{equation}
\label{eq:losstext}
  \mathcal{L}_{text} = \frac{1}{HW}|||\nabla I_{F}|-max(|\nabla I_{1}|, |\nabla I_{2}|)||_{1}
\end{equation}

\noindent where $\nabla(\cdot)$ denotes the gradient operator, and $\max(\cdot)$ selects the maximum value. Once the fused image captures adequate intensity and detailed information, structural similarity is computed between the fused image and each source image to enhance visual consistency and perceptual quality.

The SSIM-based loss is formulated as follows:
\begin{equation}
\label{eq:lossssim}
\begin{aligned}
\mathcal{L}_{SSIM} = \gamma_{1}(1&-SSIM(I_{F},I_{1})) \\ &+\gamma_{2}(1-SSIM(I_{F},I_{2}))
\end{aligned}
\end{equation}

\noindent where $SSIM(\cdot)$ denotes the structural similarity computation. The coefficients $\gamma_{1}$ and $\gamma_{2}$ weight the SSIM between each source image and the fused result. To guarantee equal contribution from both sources, both weights are set to 0.5.

\begin{figure*}[!t]
\centering
\includegraphics[width=1.0 \textwidth]{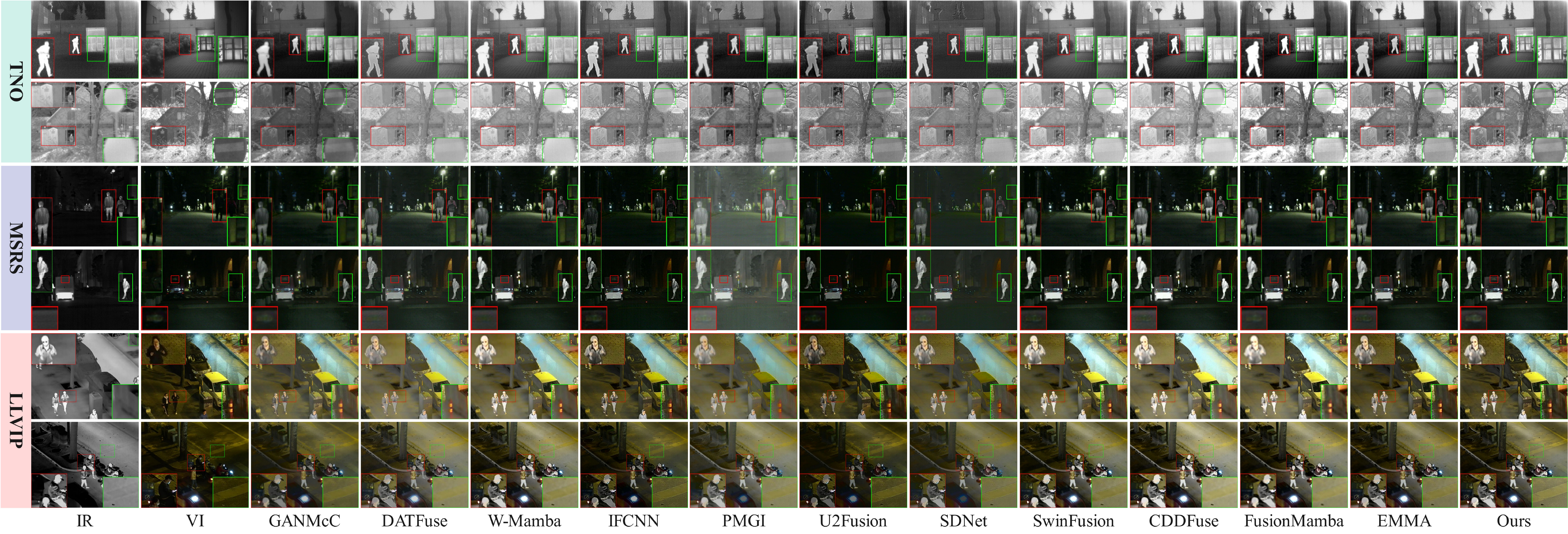}
\caption{Qualitative comparison of WIFE-Fusion and eleven benchmark methods on the TNO dataset (top two rows), MSRS dataset (middle two rows) and LLVIP dataset (bottom two rows) for infrared and visible image fusion. For better visual comparison, key regions are highlighted and enlarged using \textcolor{red}{red} and \textcolor{green}{green} bounding boxes.}
\label{Qualitative_IVIF}
\end{figure*}

\section{Experiments and Analysis} \label{sec:experiment}
Sec.~\ref{subsec:setting} details the experimental setup, including datasets, training configurations, comparative methods benchmark, and evaluation metrics. Sec.~\ref{subsec:IVIFresult}, Sec.~\ref{subsec:RGBNIRresult}, and Sec.~\ref{subsec:MIFresult} present the fusion performance of WIFE-Fusion across three tasks: infrared-visible image fusion (IVIF), RGB and near-infrared (RGBNIR) image fusion, and medical image fusion (MIF). Sec. \ref{subsec:Ablation} conducts ablation studies to evaluate the effectiveness of the network architecture and loss functions. Finally, Sec.~\ref{subsec:downstream} demonstrates comparative experiments applying the fusion algorithm to downstream object detection tasks, validating the superior performance of WIFE-Fusion in downstream applications.

\subsection{Experimental Settings} \label{subsec:setting}

\subsubsection{Datasets}
\label{subsubsec:datasets}
To thoroughly validate the effectiveness of WIFE-Fusion, we performed a comprehensive performance characterization across three multi-modal image fusion tasks such as infrared-visible image fusion (IVIF), RGB and near-infrared (RGBNIR) fusion, and medical image fusion (MIF) by using five challenging and diverse datasets. For IVIF, the model was trained on 12,025 training image pairs from the LLVIP dataset \cite{jia2021llvip}, with 50 randomly selected test pairs used for performance assessment. To assess the generalization capability of our network, we additionally conducted generalization experiments on 41 image pairs from the TNO dataset \cite{toet2017tno} and 45 pairs from the MSRS dataset \cite{tang2022piafusion}. For RGBNIR image fusion, we tested WIFE-Fusion's performance on 38 image pairs from the RGB-NIR dataset \cite{brown2011multi} using pre-trained weights from the IVIF task. For MIF, a total of 427 CT-MRI and PET-MRI image pairs from the Harvard medical dataset (\url{https://www.med.harvard.edu/AANLIB/home.html}) were used for training, and 24 CT-MRI and 24 PET-MRI image pairs were used for testing.

\subsubsection{Implementation Details}
\label{subsubsec:details}
All experiments were performed on a single NVIDIA GeForce RTX 4090D GPU. During training, image pairs from the LLVIP and Harvard Medical datasets were randomly cropped into 128×128 patches and normalized to the [0, 1] range. The Adam optimizer was used with a learning rate of $1 \times 10^{-3}$ and a batch size of 16. The model was trained for 50 epochs for IVIF tasks and 1000 epochs for MIF. To balance the contributions of the loss components in Eq. \ref{eq:Loss Function}, the hyperparameters $\alpha$, $\beta$, and $\gamma$ were set to 2, 10, and 1, respectively.

\subsubsection{Comparison methods}
\label{subsubsec:comparison}
We selected a total of 14 state-of-the-art methods for comparison, the key configuration of all comparative experiments are shown in Tab. \ref{tab:parameters}. Specifically, our comparison methods include the AutoEncoder(AE)-based method EMFusion \cite{xu2021emfusion} and CDDFuse \cite{zhao2023cddfuse}, the CNN-based methods U2Fusion \cite{xu2020u2fusion}, SDNet \cite{zhang2021sdnet}, PMGI \cite{zhang2020rethinking}, GeSeNet \cite{li2023gesenet}, and IFCNN \cite{zhang2020ifcnn}, the GAN-based method GANMcC \cite{ma2020ganmcc}, the INN-based method INet \cite{he2025inet}, the Transformer-based methods DATFuse \cite{tang2023datfuse}, SwinFusion \cite{ma2022swinfusion}, and EMMA \cite{zhao2024emma}, and the Mamba-based method FusionMamba \cite{xie2024fusionmamba} and W-Mamba \cite{zhang2025exploring}. Among them, GANMcC \cite{ma2020ganmcc}, DATFuse \cite{tang2023datfuse} and W-Mamba \cite{zhang2025exploring} are specifically designed for infrared and visible image fusion, while EMFusion \cite{xu2021emfusion}, GeSeNet \cite{li2023gesenet} and INet \cite{he2025inet} are tailored for MIF. The remaining eight methods represent advanced network frameworks applicable to general multi-modal image fusion tasks. All compared models are openly accessible, and we utilized their official pretrained weights from public releases to ensure fair performance evaluation.

\subsubsection{Evaluation Metrics}
\label{subsec:metrics}
We employ six metrics for quantitative evaluation: one information-theoretic metric—Feature Mutual Information ($FMI$) \cite{haghighat2011non,qu2002information}; two human visual perception–based metrics—Human Visual Perception ($Q_{cb}$) \cite{chen2009new} and Visual Information Fidelity ($VIF$) \cite{han2013new}; one structure-similarity metrics—Peilla’s metric ($Q_{w}$) \cite{piella2003new}; and two image feature–based metrics—Gradient-based Similarity Measurement ($Q_{abf}$) \cite{xydeas2000objective} and Phase Congruency ($Q_{p}$) \cite{zhao2007performance}. Notably, higher values for all six metrics indicate better fusion performance.

\begin{table*}[!t]
\centering
\caption{Quantitative results of WIFE‑Fusion and eleven benchmark methods are compared on the TNO, MSRS, and LLVIP datasets for infrared and visible image fusion. For each evaluation metric, the top three performers are highlighted in \textcolor{red}{red}, \textcolor{blue}{blue}, and \textcolor{green}{green}, respectively. The symbol $\uparrow$ denotes that higher scores indicate better performance.}
\resizebox{\textwidth}{!}{
\begin{tabular}{c|cccccc}
\toprule
\multirow{2}{*}{\rule{0pt}{2.5ex}\textbf{Method}} 
& \multicolumn{6}{c}{\textbf{Infrared and Visible Image Fusion(TNO/MSRS/LLVIP)}} \\
\cline{2-7}
& \rule{0pt}{2.5ex} \textbf{FMI$\uparrow$} & \textbf{Q$_{cb}\uparrow$} & \textbf{VIF$\uparrow$} & \textbf{Q$_{w}\uparrow$} & \textbf{Q$_{abf}\uparrow$} & \textbf{Q$_{p}\uparrow$}  \\
\midrule
GANMcC\cite{ma2020ganmcc}    
& 0.8962 / 0.9365 / 0.9064 & 0.4369 / 0.4589 / 0.3007 & 0.5270 / 0.6730 / 0.5681 & 0.5189 / 0.5926 / 0.4825 & 0.2884 / 0.3439 / 0.2595 & 0.2389 / 0.2476 / 0.2712 \\
DATFuse\cite{tang2023datfuse}  
& 0.8793 / 0.9426 / 0.9066 & 0.4155 / 0.4496 / 0.3335 & 0.6966 / 0.8853 / 0.7011 & 0.6483 / 0.8468 / 0.6835 & 0.5065 / 0.5608 / 0.4697 & \textcolor{green}{0.3786} / 0.3594 / 0.2951 \\
W-Mamba\cite{zhang2025exploring}  
& 0.8924 / \textcolor{green}{0.9494} / 0.9152 & 0.4551 / \textcolor{green}{0.5415} / 0.4036 & \textcolor{blue}{0.8329} / 0.9626 / 0.8620 & 0.7010 / 0.8881 / 0.7609 & \textcolor{blue}{0.5492} / \textcolor{green}{0.6017} / 0.5498 & \textcolor{blue}{0.4297} / \textcolor{blue}{0.4335} / 0.3604 \\
IFCNN\cite{zhang2020ifcnn}     
& 0.8949 / 0.9425 / 0.9092 & 0.4890 / 0.4878 / \textcolor{green}{0.4501} & 0.6370 / 0.6332 / 0.7096 & \textcolor{blue}{0.7454} / 0.8245 / 0.8488 & 0.4765 / 0.4550 / 0.5906 & 0.3060 / 0.2455 / \textcolor{green}{0.3761} \\
PMGI\cite{zhang2020rethinking}   
& 0.8994 / 0.9292 / 0.9072 & 0.4761 / 0.2231 / 0.2949 & 0.6418 / 0.7394 / 0.6871 & 0.6895 / 0.6229 / 0.6173 & 0.4122 / 0.3719 / 0.4187 & 0.2746 / 0.2403 / 0.2438 \\
U2Fusion\cite{xu2020u2fusion} 
& 0.8893 / 0.9427 / 0.9043 & \textcolor{blue}{0.5102} / 0.4465 / \textcolor{blue}{0.4555} & 0.6273 / 0.4968 / 0.6487 & 0.7214 / 0.7450 / 0.7370 & 0.4254 / 0.3403 / 0.4957 & 0.2862 / 0.2169 / 0.3565 \\
SDNet\cite{zhang2021sdnet}
& 0.8972 / 0.9354 / 0.9089 & 0.4698 / 0.3162 / 0.3357 & 0.5916 / 0.5213 / 0.6514 & 0.7286 / 0.6995 / 0.7771 & 0.4318 / 0.3514 / 0.5328 & 0.2774 / 0.2070 / 0.2843 \\
SwinFusion\cite{ma2022swinfusion}
& \textcolor{green}{0.9025} / \textcolor{blue}{0.9519} / \textcolor{blue}{0.9197} & 0.4855 / \textcolor{red}{0.5537} / 0.4215 & 0.7626 / \textcolor{green}{0.9653} / \textcolor{green}{0.8808} & 0.7292 / \textcolor{blue}{0.8978} / \textcolor{blue}{0.8852} & \textcolor{green}{0.5201} / 0.5747 / \textcolor{blue}{0.6407} & 0.3583 / 0.4109 / 0.3518 \\
CDDFuse\cite{zhao2023cddfuse}
& \textcolor{blue}{0.9057} / 0.9484 / \textcolor{green}{0.9188} & 0.4505 / 0.5347 / 0.3958 & \textcolor{green}{0.7821} / \textcolor{blue}{1.0447} / \textcolor{blue}{0.9542} & \textcolor{green}{0.7443} / 0.8905 / 0.8404 & 0.5147 / \textcolor{blue}{0.6111} / 0.6149 & 0.3674 / \textcolor{green}{0.4216} / \textcolor{blue}{0.4170} \\
FusionMamba\cite{xie2024fusionmamba}
& 0.8983 / 0.9411 / 0.9037 & 0.4688 / 0.5172 / 0.3703 & 0.5601 / 0.7416 / 0.5581 & 0.5854 / 0.7605 / 0.2130 & 0.3374 / 0.4187 / 0.2031 & 0.2008 / 0.2228 / 0.0801 \\
EMMA\cite{zhao2024emma}
& 0.8957 / 0.9492 / 0.9148 & \textcolor{green}{0.4913} / 0.5274 / 0.4282 & 0.7085 / 0.9501 / 0.8232 & 0.7126 / \textcolor{green}{0.8943} / \textcolor{green}{0.8811} & 0.4584 / 0.5712 / \textcolor{green}{0.6179} & 0.3300 / 0.3934 / 0.3478 \\
\textbf{$\star$Ours}     
& \textcolor{red}{0.9160} / \textcolor{red}{0.9529} / \textcolor{red}{0.9273} & \textcolor{red}{0.5217} / \textcolor{blue}{0.5430} / \textcolor{red}{0.4706} & \textcolor{red}{0.8825} / \textcolor{red}{1.0607} / \textcolor{red}{1.0416} & \textcolor{red}{0.8350} / \textcolor{red}{0.9162} / \textcolor{red}{0.9249} & \textcolor{red}{0.6343} / \textcolor{red}{0.6542} / \textcolor{red}{0.7423} & \textcolor{red}{0.4565} / \textcolor{red}{0.4677} / \textcolor{red}{0.5098}  \\
\bottomrule
\end{tabular}
}
\label{tab:Quantitative_IVIF}
\end{table*}

\begin{figure}[!t]
\centering
\includegraphics[width=1.0 \columnwidth]{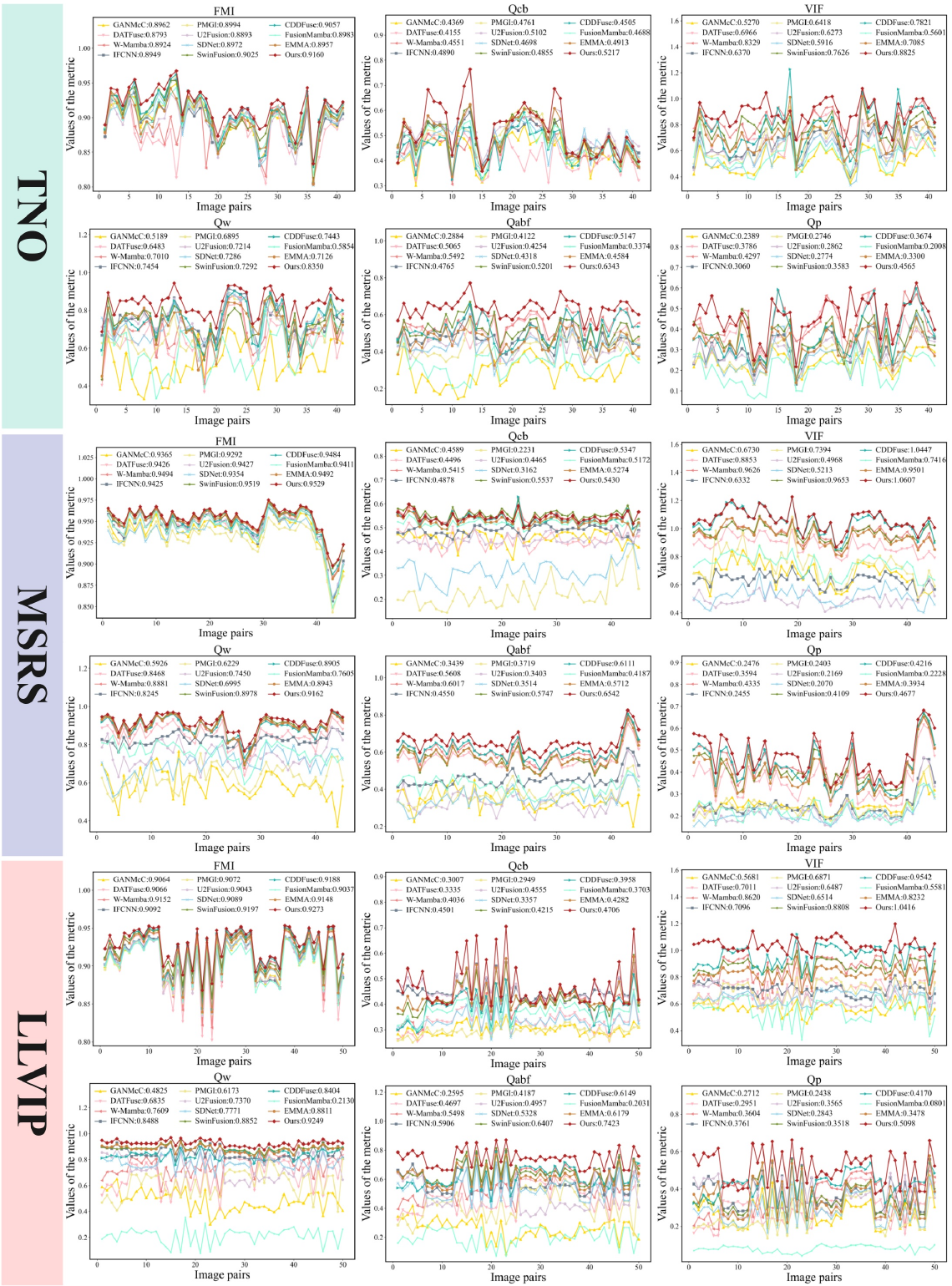}
\caption{Quantitative comparison of WIFE-Fusion and eleven benchmark methods on the TNO, MSRS, and LLVIP datasets for infrared and visible image fusion. Our approach is shown as the \textcolor{red}{red} line, with each legend displaying the method’s mean scores. For all six evaluation metrics, higher values denote better performance.}
\label{fig:Quantitative_IVIF}
\end{figure}

\begin{figure}[!t]
\centering
\includegraphics[width=1.0 \columnwidth]{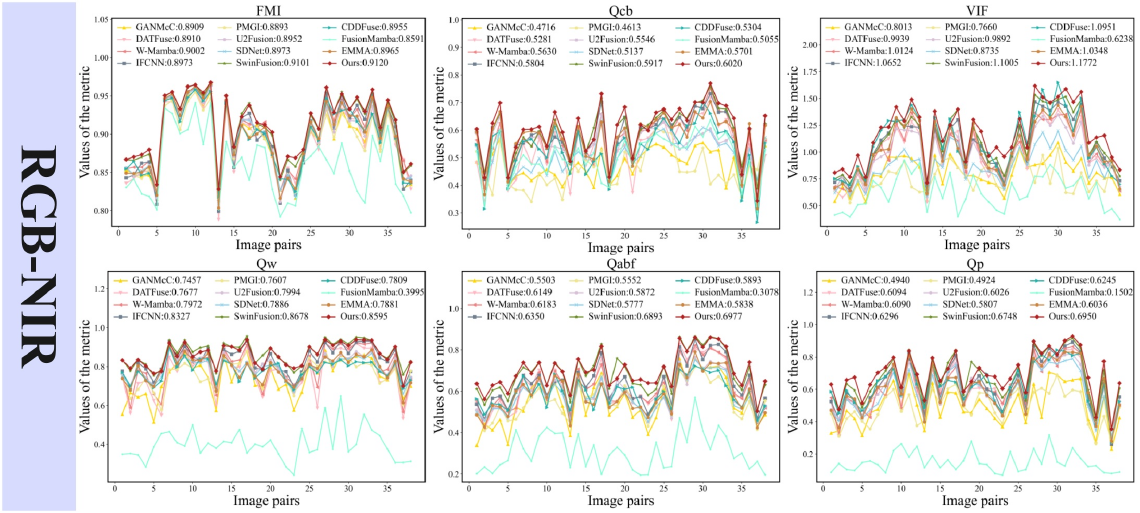}
\caption{Quantitative comparison of WIFE-Fusion and eleven benchmark methods on the RGB-NIR dataset for RGB and near-infrared image fusion. Our approach is shown as the \textcolor{red}{red} line, with each legend displaying the method’s mean scores. For all six evaluation metrics, higher values denote better performance.}
\label{fig:Quantitative_RGBNIR}
\end{figure}

\begin{table}[!t]
\centering
\caption{Quantitative results of WIFE‑Fusion and eleven benchmark methods are compared on the RGB-NIR dataset for RGB and near-infrared image fusion. For each evaluation metric, the top three performers are highlighted in \textcolor{red}{red}, \textcolor{blue}{blue}, and \textcolor{green}{green}, respectively. The symbol $\uparrow$ denotes that higher scores indicate better performance.
}
\resizebox{\columnwidth}{!}{
\begin{tabular}{c|cccccc}
\toprule
\multirow{2}{*}{\rule{0pt}{2.5ex}\textbf{Method}} 
& \multicolumn{6}{c}{\textbf{RGB and Near-Infrared Image Fusion(RGB-NIR)}} \\
\cline{2-7}
&\rule{0pt}{2.5ex} \textbf{FMI$\uparrow$} & \textbf{Q$_{cb}\uparrow$} & \textbf{VIF$\uparrow$} & \textbf{Q$_{w}\uparrow$} & \textbf{Q$_{abf}\uparrow$} & \textbf{Q$_{p}\uparrow$} \\
\midrule
GANMcC\cite{ma2020ganmcc}    
& 0.8909 & 0.4716 & 0.8013 & 0.7457 & 0.5503 & 0.4940  \\
DATFuse\cite{tang2023datfuse}  
& 0.8910 & 0.5281 & 0.9939 & 0.7677 & 0.6149 & 0.6094  \\
W-Mamba\cite{zhang2025exploring}  
& \textcolor{green}{0.9002} & 0.5630 & 1.0124 & 0.7972 & 0.6183 & 0.6090 \\
IFCNN\cite{zhang2020ifcnn}     
& 0.8973 & \textcolor{green}{0.5804} & 1.0652 & \textcolor{green}{0.8327} & \textcolor{green}{0.6350} & \textcolor{green}{0.6296} \\
PMGI\cite{zhang2020rethinking}   
& 0.8893 & 0.4613 & 0.7660 & 0.7607 & 0.5552 & 0.4924 \\
U2Fusion\cite{xu2020u2fusion} 
& 0.8952 & 0.5546 & 0.9892 & 0.7994 & 0.5872 & 0.6026 \\
SDNet\cite{zhang2021sdnet}
& 0.8973 & 0.5137 & 0.8735 & 0.7886 & 0.5777 & 0.5807 \\
SwinFusion\cite{ma2022swinfusion}
& \textcolor{blue}{0.9101} & \textcolor{blue}{0.5917} & \textcolor{blue}{1.1005} & \textcolor{red}{0.8678} & \textcolor{blue}{0.6893} & \textcolor{blue}{0.6748} \\
CDDFuse\cite{zhao2023cddfuse}
& 0.8955 & 0.5304 & \textcolor{green}{1.0951} & 0.7809 & 0.5893 & 0.6245\\
FusionMamba\cite{xie2024fusionmamba}
& 0.8591 & 0.5055 & 0.6238 & 0.3995 & 0.3078 & 0.1502 \\
EMMA\cite{zhao2024emma}
& 0.8965 & 0.5701 & 1.0348 & 0.7881 & 0.5838 & 0.6036 \\
\textbf{$\star$Ours}     
& \textcolor{red}{0.9120} & \textcolor{red}{0.6020} & \textcolor{red}{1.1772} & \textcolor{blue}{0.8595} & \textcolor{red}{0.6977} & \textcolor{red}{0.6950} \\
\bottomrule
\end{tabular}
}
\label{tab:Quantitative_RGBNIR}
\end{table}

\begin{figure*}[!t]
\centering
\includegraphics[width=1.0 \textwidth]{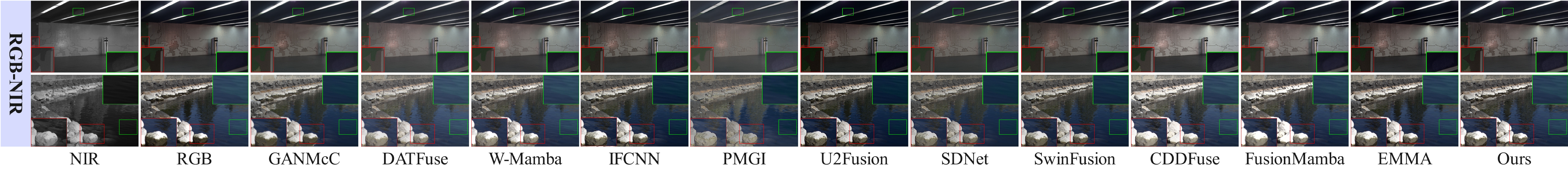}
\caption{Qualitative comparison of WIFE-Fusion and eleven benchmark methods on the RGB-NIR dataset for RGB and near-infrared image fusion. For better visual comparison, key regions are highlighted and enlarged using \textcolor{red}{red} and \textcolor{green}{green} bounding boxes.}
\label{fig:Qualitative_RGBNIR}
\end{figure*}

\begin{figure*}[!t]
\centering
\includegraphics[width=1.0 \textwidth]{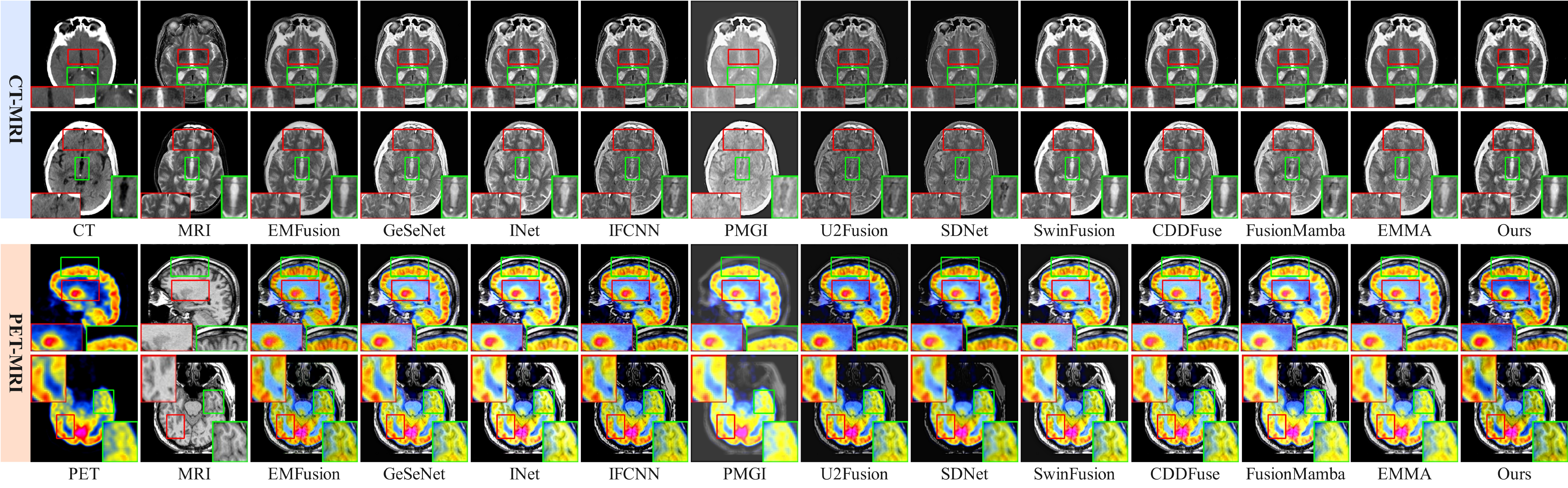}
\caption{Qualitative comparison of WIFE-Fusion and eleven benchmark methods on the Harvard medical image dataset for CT-MRI image fusion (top two rows) and PET-MRI image fusion (bottom two rows). For better visual comparison, key regions are highlighted and enlarged using \textcolor{red}{red} and \textcolor{green}{green} bounding boxes.}
\label{fig:Qualitative_MIF}
\end{figure*}

\subsection{Experimental Results on Infrared and Visible Image Fusion} \label{subsec:IVIFresult}
\subsubsection{Qualitative Analysis}
In Fig.~\ref{Qualitative_IVIF}, we present a qualitative assessment of infrared-visible fusion (IVIF) results on the TNO, MSRS, and LLVIP datasets to visually demonstrate the dominance of WIFE‑Fusion. Key regions are highlighted with red and green boxes for direct comparative analysis. The compared methods have shortcomings such as,  \textbf{Target Preservation Deficits}: GANMcC and IFCNN fail to retain prominent infrared targets (green box in the fourth row and red box in the fifth row); \textbf{Edge Information Degradation}: DATFuse, W-Mamba, PMGI and CDDFuse exhibit insufficient emphasis on visible-light edge details, leading to blurred or absent edges (green boxes in the fifth/sixth rows, LLVIP); \textbf{Intensity Fidelity Issues}: U2Fusion struggles to maintain source-image intensity, producing fused outputs with dimmed targets and flattened brightness (MSRS and LLVIP results); \textbf{Contrast Loss}: SDNet fails to inherit visible‑light intensity, yielding low‑contrast fusions (first and third rows); \textbf{Contrast and Detail Degradation}: W-Mamba, SwinFusion, CDDFuse and EMMA compromise visible-light contrast, leading to attenuated or lost information (green boxes in the second, fifth, and sixth rows, and red box in the fourth row); \textbf{Detail Blurring}: FusionMamba generates blurred outputs with substantial detail loss, such as indistinct floor tiles (red box, fifth row) and obscured crate text (red box, sixth row). In contrast, our network fully leverages complementary source-image information, producing fusions with clear visual hierarchy, robust contrast, rich texture detail, and alignment with human visual perception.

\subsubsection{Quantitative Analysis}
The quantitative comparison of our network for the IVIF task is presented in Fig.~\ref{fig:Quantitative_IVIF} and Tab.~\ref{tab:Quantitative_IVIF}. Significantly, our method outperforms all baseline methods across all metrics, expect for $Q_{cb}$ on the MSRS dataset, where SwinFusion achieves a slightly higher score. This demonstrates that WIFE-Fusion produces fusions with superior feature retention, enhanced structural similarity (SSIM), and improved visual perceptual quality (VIF).

\subsection{Experimental Results on RGB and Near-Infrared Image Fusion} \label{subsec:RGBNIRresult}

\subsubsection{Qualitative Analysis}
In Fig.~\ref{fig:Qualitative_RGBNIR}, we present a qualitative evaluation of RGBNIR fusion results on the RGB-NIR dataset to visually demonstrate the superior image quality of WIFE-Fusion. Key regions are highlighted with red and green boxes for straightforward comparative analysis. The compared methods have shortcoming such as, \textbf{Contrast and Edge Degradation}: GANMcC and IFCNN fail to maintain RGB source contrast, leading to lost edge details (e.g., the waterline boundary in the green box of Row 2). \textbf{NIR Intensity Over-amplification}: DATFuse, W-Mamba and CDDFuse over-amplifies NIR intensity, causing unnatural color distortions in regions like the tree-trunk hue (red box, Row 1) and stone water stains (red box, Row 2). \textbf{Low-Contrast Artifact}: PMGI and SDNet generate low-contrast fusions with weakened edge and intensity information (e.g., stone water stains in Row 2 red box; waterline boundary in Row 2 green box). \textbf{Modality Balance Deficiency}: U2Fusion, SwinFusion, and EMMA struggle to balance cross-modality contrast, resulting in faded or absent details (e.g., trunk-leaf boundary in Row 1 red box; white texture in Row 1 green box). \textbf{Global Blurring Artifact}: FusionMamba produces generally blurred images, as seen in the red boxes of both the first and second rows. In contrast, our method fully leverages complementary edge and color cues from RGB and NIR modalities, yielding fusions with excellent color fidelity, sharp edge definition, and rich textural details.

\subsubsection{Quantitative Analysis}
The quantitative comparison of our network for the RGBNIR fusion is presented in Fig.~\ref{fig:Quantitative_RGBNIR} and Tab.~\ref{tab:Quantitative_RGBNIR}. Apart from the $Q_{w}$ metric on the RGB-NIR dataset, where SwinFusion achieves a marginally higher score, the performance of the remaining five metrics all achieved leading results. This confirms that WIFE-Fusion generates fused images with superior overall quality, including enhanced structural and spectral consistency.

\begin{table*}[!t]
\centering
\caption{
Quantitative results of WIFE‑Fusion and eleven benchmark methods are compared on the Harvard Medical Image dataset for CT-MRI and PET-MRI image fusion. For each evaluation metric, the top three performers are highlighted in \textcolor{red}{red}, \textcolor{blue}{blue}, and \textcolor{green}{green}, respectively. The symbol $\uparrow$ denotes that higher scores indicate better performance.}
\resizebox{\textwidth}{!}{
\begin{tabular}{c|cccccc}
\toprule
\multirow{2}{*}{\rule{0pt}{2.5ex}\textbf{Method}} 
& \multicolumn{6}{c}{\textbf{Medical Image Fusion(CT-MRI/PET-MRI)}} \\
\cline{2-7}
& \rule{0pt}{2.5ex} \textbf{FMI$\uparrow$} & \textbf{Q$_{cb}\uparrow$} & \textbf{VIF$\uparrow$} & \textbf{Q$_{w}\uparrow$} & \textbf{Q$_{abf}\uparrow$} & \textbf{Q$_{p}\uparrow$}  \\
\midrule
EMFusion\cite{xu2021emfusion}    
& 0.8687 / 0.8393 & \textcolor{green}{0.6642} / 0.5916 & \textcolor{green}{0.5523} / 0.6226 & 0.5311 / 0.8546 & 0.4751 / 0.6439 & \textcolor{red}{0.4363} / 0.4723 \\
GeSeNet\cite{li2023gesenet}  
& 0.8763 / 0.8441 & 0.3672 / 0.3964 & 0.5076 / 0.6690 & \textcolor{blue}{0.7683} / \textcolor{green}{0.8838} & \textcolor{blue}{0.6290} / \textcolor{red}{0.7002} & 0.3830 / \textcolor{blue}{0.5066} \\
INet\cite{he2025inet}
& 0.8640 / 0.8454 & 0.6200 / 0.5506 & 0.4648 / 0.6666 & 0.6982 / 0.8757 & 0.5840 / 0.6781 & 0.3819 / 0.4770 \\
IFCNN\cite{zhang2020ifcnn}     
& 0.8629 / 0.8162 & \textcolor{blue}{0.6707} / \textcolor{green}{0.5989} & 0.4063 / 0.4948 & 0.7218 / 0.7979 & 0.5648 / 0.5847 & 0.2998 / 0.3934 \\
PMGI\cite{zhang2020rethinking}   
& \textcolor{red}{0.8890} / 0.8307 & 0.2250 / 0.1804 & 0.4271 / 0.4632 & 0.7345 / 0.1270 & 0.4535 / 0.2014 & 0.2712 / 0.1018 \\
U2Fusion\cite{xu2020u2fusion} 
& 0.8592 / 0.8130 & 0.3209 / 0.3507 & 0.3373 / 0.5108 & 0.5972 / 0.6571 & 0.4582 / 0.4681 & 0.2680 / 0.3981 \\
SDNet\cite{zhang2021sdnet}
& 0.8625 / 0.8074 & 0.3082 / 0.3502 & 0.3386 / 0.4463 & 0.6194 / 0.3745 & 0.4291 / 0.2939 & 0.2627 / 0.2736 \\
SwinFusion\cite{ma2022swinfusion}
& \textcolor{blue}{0.8804} / \textcolor{blue}{0.8510} & 0.6607 / 0.3514 & \textcolor{blue}{0.5578} / \textcolor{green}{0.7123} & \textcolor{green}{0.7617} / \textcolor{blue}{0.8850} & 0.5820 / \textcolor{green}{0.6912} & \textcolor{green}{0.3867} / \textcolor{green}{0.4960} \\
CDDFuse\cite{zhao2023cddfuse}
& 0.8734 / \textcolor{green}{0.8485} & 0.6625 / \textcolor{blue}{0.6004} & 0.5256 / \textcolor{blue}{0.7123} & 0.7574 / 0.8802 & \textcolor{green}{0.6164} / 0.6808 & 0.3775 / 0.4900 \\
FusionMamba\cite{xie2024fusionmamba}
& 0.8690 / 0.8439 & 0.6399 / 0.5702 & 0.4060 / 0.6206 & 0.7225 / 0.8688 & 0.5543 / 0.6511 & 0.2982 / 0.4595 \\
EMMA\cite{zhao2024emma}
& 0.8739 / 0.8259 & 0.5557 / 0.4804 & 0.4643 / 0.5799 & 0.7333 / 0.8138 & 0.5003 / 0.5900 & 0.3072 / 0.4260 \\
\textbf{$\star$Ours}     
& \textcolor{green}{0.8801} / \textcolor{red}{0.8524} & \textcolor{red}{0.6963} / \textcolor{red}{0.6351} & \textcolor{red}{0.5780} / \textcolor{red}{0.7649} & \textcolor{red}{0.7912} / \textcolor{red}{0.8856} & \textcolor{red}{0.6365} / \textcolor{blue}{0.6946} & \textcolor{blue}{0.4306} / \textcolor{red}{0.5127} \\
\bottomrule
\end{tabular}
}
\label{tab:Quantitative_MIF}
\end{table*}

\begin{figure}[!t]
\centering
\includegraphics[width=1.0 \columnwidth]{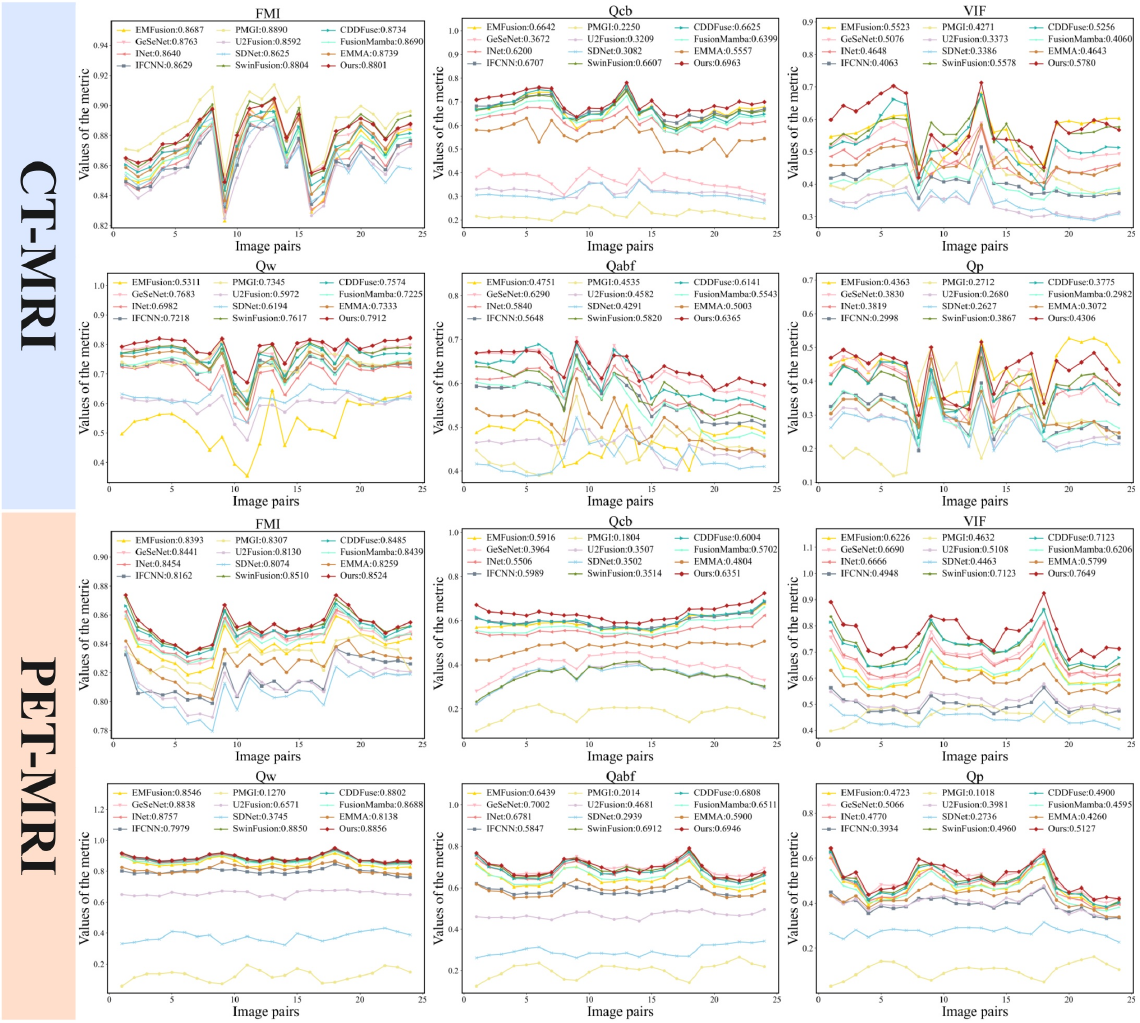}
\caption{Quantitative comparison of WIFE-Fusion and eleven benchmark methods on the Harvard Medical Image dataset for CT‑MRI and PET‑MRI image fusion. Our approach is shown as the \textcolor{red}{red} line, with each legend displaying the method’s mean scores. For all six evaluation metrics, higher values denote better performance.
}
\label{fig:Quantitative_MIF}
\end{figure}

\subsection{Experimental Results on Medical Image Fusion}\
\label{subsec:MIFresult}
\subsubsection{Qualitative Analysis}
In Fig.~\ref{fig:Qualitative_MIF}, we present a qualitative analysis of CT–MRI and PET–MRI fusion results on the Harvard medical dataset to visually demonstrate the superiority of our fusion network. Key regions are highlighted with red and green boxes for direct visual comparison. 

For CT-MRI image fusion, other comprasion methods have shortcomings such as, \textbf{CT Feature Preservation Failure}: EMFusion fails to retain bright CT features (e.g., highlights in green boxes, Row 1–2). \textbf{PET Structure Contrast attenuation}: GeSeNet, SwinFusion, CDDFuse and EMMA suffer from contrast attenuation in MRI-derived structural details (e.g., soft tissue boundaries in red boxes, Row 1 and 2). \textbf{Fine MRI Detail Loss}: PMGI fails to preserve fine-grained MRI textures (e.g., tissue textures in green box in Row 1 and red box in Row 2). \textbf{Contrast Inversion Challenges}: INet, IFCNN, U2Fusion, SDNet, and FusionMamba struggle with regions of opposing contrast, leading to combined loss of structural and intensity information (e.g., margins in green box, Row 2). 

In PET–MRI image fusion, other comprasion methods have shortcomings such as, \textbf{PET Intensity Degradation}: All eleven baseline methods exhibit weakened or absent PET metabolic signal (e.g., FDG uptake patterns in red boxes, Row 1–2). \textbf{MRI Structural Detail Loss}: GeSeNet, IFCNN, PMGI, U2Fusion, SDNet, CDDFuse, FusionMamba and EMMA similarly struggle to preserve critical MRI detail information (green boxes in the first and second rows). In contrast, our network effectively fuses cross-modality features to generate outputs with rich anatomical details, faithful intensity preservation, and robust structural consistency, aligning with clinical diagnostic requirements.

\subsubsection{Quantitative Analysis}
The quantitative comparison for MIF is presented in Fig.~\ref{fig:Quantitative_MIF} and Tab.~\ref{tab:Quantitative_MIF}. While our $FMI$ for CT–MRI image fusion is slightly lower than that of PMGI and SwinFusion, and our $Q_{p}$ in CT–MRI image fusion and $Q_{abf}$ in PET–MRI image fusion rank second among all methods, WIFE-Fusion dominates the remaining metrics across both tasks. This demonstrates that our method produces medical fusions with superior information retention, robust structural consistency, and clinical-grade visual quality, aligning with diagnostic requirements.

\begin{figure*}[!t]
\centering
\includegraphics[width=1.0 \textwidth]{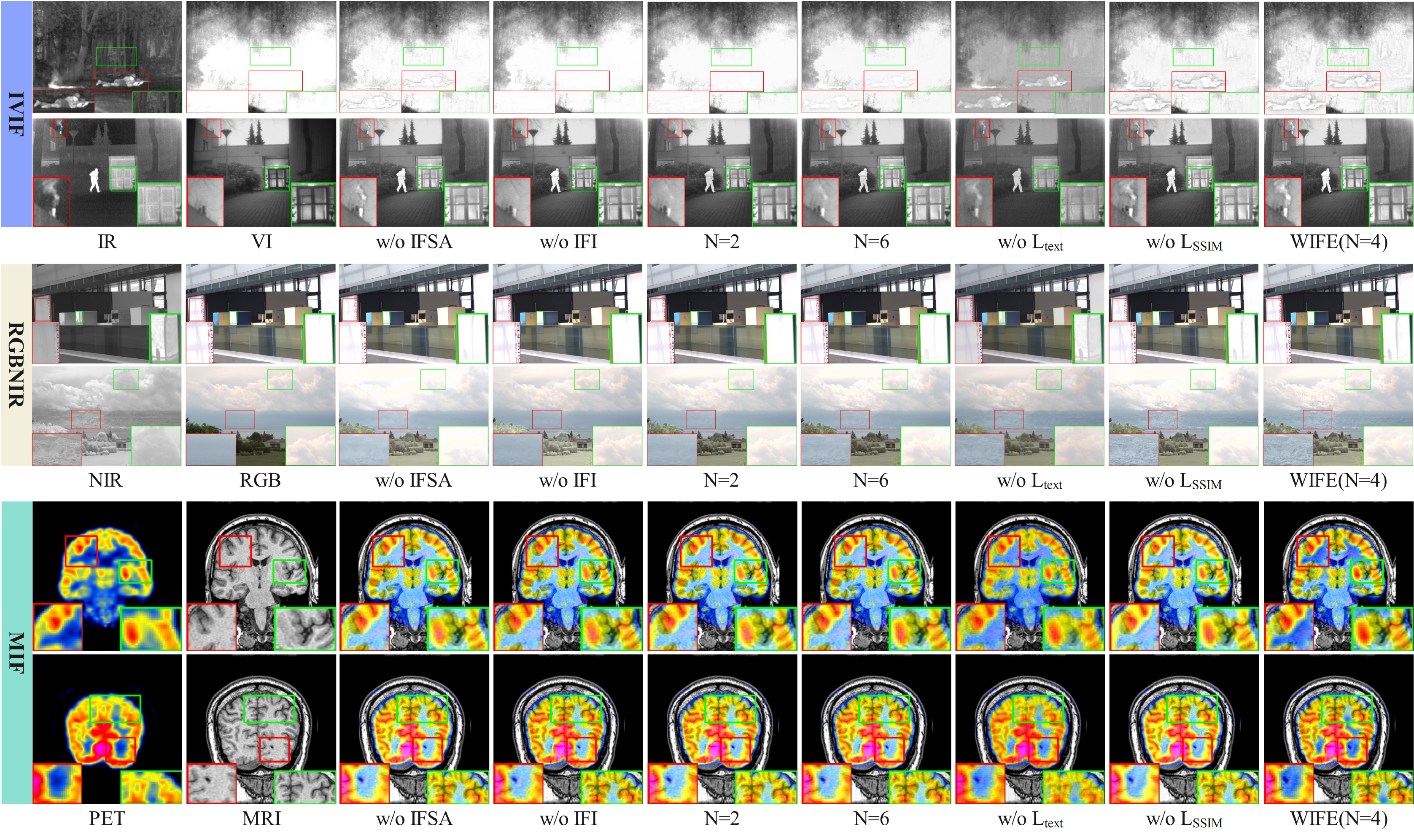}
\caption{Qualitative comparison of ablation studies on the network architecture and loss function across three fusion tasks: Infrared and Visible Image Fusion (IVIF), RGB and Near-Infrared Image Fusion (RGBNIR), and Medical Image Fusion (MIF). Key regions in the images are highlighted with \textcolor{red}{red} and \textcolor{green}{green} boxes and enlarged to facilitate visual comparison. $N$ denotes the number of sequential WIFE modules in the network, which is set to 4 in WIFE-Fusion.}
\label{fig:Qualitative_Ablation}
\end{figure*}

\begin{table*}[!t]
\centering
\caption{Quantitative ablation results on network architecture and loss function are reported across three fusion tasks: IVIF, RGBNIR, and MIF. The best scores for each metric are marked in \textcolor{red}{red}. Here, $N$ denotes the number of consecutive WIFE modules in the architecture, set to 4 in the full WIFE-Fusion model. The symbol $\uparrow$ indicates that higher values reflect improved performance.}
\resizebox{\textwidth}{!}{
\begin{tabular}{c|cccccc}
\toprule
\multirow{2}{*}{} 
& \multicolumn{6}{c}{\textbf{Ablation Study(IVIF / RGBNIR / MIF)}} \\
\cline{2-7}
& \rule{0pt}{2.5ex} \textbf{FMI$\uparrow$} & \textbf{Q$_{cb}\uparrow$} & \textbf{VIF$\uparrow$} & \textbf{Q$_{w}\uparrow$} & \textbf{Q$_{abf}\uparrow$} & \textbf{Q$_{p}\uparrow$}  \\
\midrule
w/o IFSA    
& 0.9138 / 0.9082 / 0.8513 & 0.4912 / 0.5748 / 0.5705 & 0.8383 / 1.1297 / 0.7354 & 0.7930 / 0.8274 / 0.8808 & 0.5838 / 0.6597 / 0.6877 & 0.4186 / 0.6675 / 0.5084 \\
w/o IFI  
& 0.9139 / 0.9105 / 0.8506 & 0.5033 / 0.5925 / 0.6114 & 0.8391 / 1.1673 / 0.7161 & 0.7650 / 0.8379 / 0.8784 & 0.5687 / 0.6705 / 0.6814 & 0.4092 / 0.6752 / 0.5031  \\
N = 2    
& 0.9133 / 0.9089 / 0.8513 & 0.5102 / 0.5891 / 0.6068 & 0.8388 / 1.1569 / 0.7280 & 0.7345 / 0.8274 / 0.8805 & 0.5565 / 0.6580 / 0.6849 & 0.4037 / 0.6664 / 0.5070 \\
N = 6   
& 0.9119 / 0.9092 / 0.8518 & 0.5065 / 0.5933 / 0.6107 & 0.8154 / 1.1661 / 0.7375 & 0.7745 / 0.8377 / 0.8805 & 0.5616 / 0.6653 / 0.6881 & 0.4020 / 0.6741 / 0.5124 \\
w/o \(L_{text}\) 
& 0.9058 / 0.9026 / 0.8518 & 0.4692 / 0.5686 / 0.6055 & 0.6139 / 1.0592 / 0.7402 & 0.6403 / 0.8186 / 0.8786 & 0.4071 / 0.6384 / 0.6819 & 0.3222 / 0.6217 / 0.5092 \\
w/o \(L_{SSIM}\)
& 0.9104 / 0.9064 / 0.8518 & 0.5131 / 0.5879 / 0.6042 & 0.7983 / 1.1383 / 0.7426 & 0.8044 / 0.8405 / 0.8854 & 0.5840 / 0.6645 / \textcolor{red}{0.6974} & 0.4102 / 0.6776 / 0.5109 \\
WIFE-Fusion
& \textcolor{red}{0.9160} / \textcolor{red}{0.9120} / \textcolor{red}{0.8524} & \textcolor{red}{0.5217} / \textcolor{red}{0.6020} / \textcolor{red}{0.6351} & \textcolor{red}{0.8825} / \textcolor{red}{1.1772} / \textcolor{red}{0.7649} & \textcolor{red}{0.8350} / \textcolor{red}{0.8595} / \textcolor{red}{0.8856} & \textcolor{red}{0.6343} / \textcolor{red}{0.6977} / 0.6946 & \textcolor{red}{0.4565} / \textcolor{red}{0.6950} / \textcolor{red}{0.5127} \\
\bottomrule
\end{tabular}
}
\label{tab:Quantitative_Ablation}
\end{table*}

\begin{figure*}[!t]
\centering
\includegraphics[width=1.0 \textwidth]{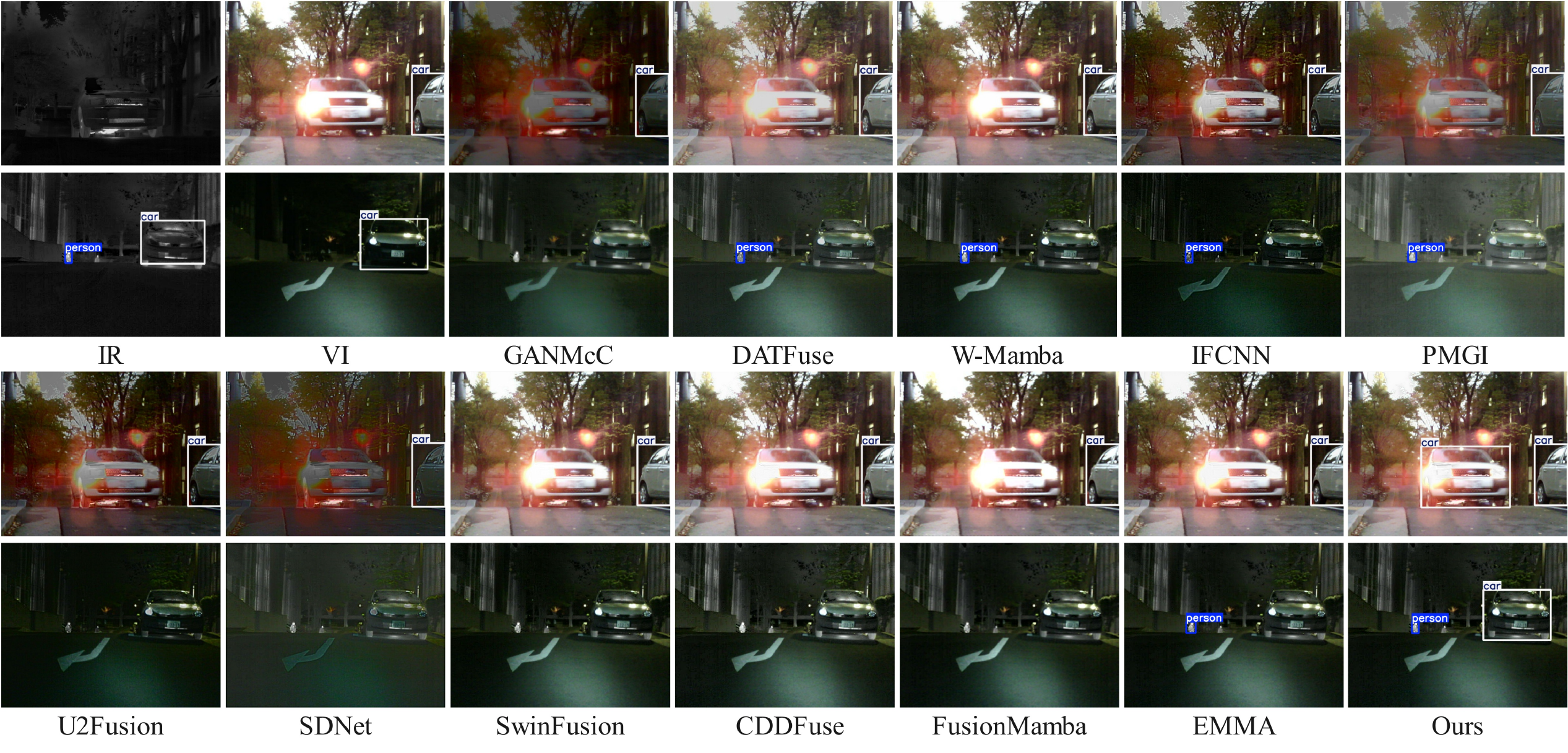}
\caption{
Qualitative comparison of WIFE-Fusion and eleven benchmark methods for object detection on the MSRS \cite{tang2022piafusion} dataset.}
\label{Qualitative_Detection}
\end{figure*}

\begin{table}
\caption{Quantitative evaluation of WIFE-Fusion against eleven benchmark methods for object detection on the MSRS \cite{tang2022piafusion} dataset, with the top performer marked in \textcolor{red}{red}. $\uparrow$ denotes that higher values indicate superior performance.
}
\footnotesize
\label{tab:Quantitative_Detection}
\footnotesize
\centering
\resizebox{\columnwidth}{!}{
    \begin{tabular}{ccccc}
    \toprule
        Method & \textbf{mAP@0.65} $\uparrow$& \textbf{mAP@0.85} $\uparrow$& \textbf{mAP@[0.5,0.95] $\uparrow$}\\
    \midrule
    IR & 0.758 & 0.420 & 0.588 \\
    VI & 0.743 & 0.355 & 0.553 \\
    GANMcC\cite{ma2020ganmcc} & 0.842 & 0.534 & 0.656 \\
    DATFuse\cite{tang2023datfuse} & 0.843 & 0.497 & 0.645 \\
    W-Mamba\cite{zhang2025exploring} & 0.848 & 0.565 & 0.659 \\
    IFCNN\cite{zhang2020ifcnn} & 0.806 & 0.473 & 0.621\\
    PMGI\cite{zhang2020rethinking} & 0.824 & 0.445 & 0.613\\
    U2Fusion\cite{xu2020u2fusion} & 0.837 & 0.498 & 0.636 \\
    SDNet\cite{zhang2021sdnet} & 0.795 & 0.487 & 0.606\\
    SwinFusion\cite{ma2022swinfusion} & 0.830 & 0.477 & 0.637\\
    CDDFuse\cite{zhao2023cddfuse} & 0.853 & 0.511 & 0.644 \\
    FusionMamba\cite{xie2024fusionmamba} & 0.797 & 0.497 & 0.622\\
    EMMA\cite{zhao2024emma} & 0.842 & 0.491 & 0.629\\
    \rowcolor[rgb]{0.9,0.9,0.9}$\star$\textbf{Ours} & \textcolor{red}{0.859} & \textcolor{red}{0.567} & \textcolor{red}{0.663}\\
    \bottomrule
    \end{tabular}
}
\end{table}

\subsection{Ablation Study} \label{subsec:Ablation}
\subsubsection{Qualitative Analysis}

We conducted ablation studies on both network architecture and loss functions to validate the core design components of WIFE‑Fusion. In Fig.~\ref{fig:Qualitative_Ablation}, $N$ denotes the number of stacked Wavelet-aware Intra-inter Frequency Enhancement (WIFE) modules (fixed at 4 in the final architecture).

\textbf{Network Architecture Ablation:} Removing the Intra‑\allowbreak Frequency Self‑Attention (IFSA) or Inter‑Frequency Interaction (IFI) components, in the IVIF task, Structural preservation is severely compromised, as evidenced by fragmented object boundaries (red boxes, Rows 1–2) and blurred texture details (green box, Row 1). Similarly, in the RGBNIR fusion task, edge extraction from NIR images is insufficient (green box in the first row), leading to lost source images details (red and green boxes in the second row). For medical image fusion, PET intensity and color distributions are not properly preserved (red and green boxes in the first row; red box in the second row). These results indicate that modifying the WIFE architecture disrupts cross-modal and cross-frequency interactions, causing structural, intensity, and detail degradation in fused outputs. This validates the indispensable role of the full WIFE design in enabling complementary feature integration.

\textbf{Loss Function Ablation:} Removing the texture loss leads to fused results that lack important structural and detail information across all tasks (red box in the second row of the IVIF task; red and green boxes in the second row of the RGBNIR task; green boxes in both rows of the medical fusion task). Without the joint optimization of texture and intensity losses, some fusion outputs in the IVIF and RGBNIR tasks exhibit inappropriate intensity levels. Omitting the SSIM loss causes low‑contrast regions and attenuated or missing intensity and texture details in all tasks (green boxes in the first and second rows of IVIF; green box in the second row of RGB–NIR; red boxes in both rows of the MIF task), degrading visual quality. Collectively, these findings demonstrate that each loss component is essential for maintaining information integrity and visual quality, validating the synergistic design of our multi-component loss function.

\subsubsection{Quantitative Analysis}

Ablation study results are presented in Tab.~\ref{tab:Quantitative_Ablation}. Most architectural modifications lead to significant degradation in quantitative metrics. This suggests that altering cross-modal or frequency-domain interaction mechanisms adversely impacts feature integration, leading to performance deterioration. The dramatic metric decline observed after removing the texture loss highlights its indispensable role in preserving structural and textural details across fusion tasks. While $Q_{abf}$ improves marginally for the MIF task when eliminating the SSIM loss, this exception is outweighed by broad performance drops in other scenarios, confirming the critical contribution of SSIM loss to perceptual quality. These findings align with qualitative comparisons, proving that architectural/loss function modifications degrade fusion performance, thereby validating the overall rationality of WIFE-Fusion's design.

\subsection{Downstream Application} \label{subsec:downstream}

\subsubsection{Qualitative Analysis}

Qualitative comparison results of applying our image fusion algorithm to downstream object detection are shown in Fig.~\ref{Qualitative_Detection}. Experimental results demonstrate that WIFE-Fusion effectively alleviates detection performance limitations caused by single-modality constraints in complex scenarios (e.g., failed detection in the first row of infrared images, incomplete target framing in the second row of infrared images, and missed detections in visible images across rows 1-2). These improvements are enabled by efficient inter-modal and frequency-domain feature interactions, which enhance cross-modality information integration. Compared to other fusion methods prone to detection omissions or false negatives, our approach delivers more robust target recognition and superior generalization in downstream applications.

\subsubsection{Quantitative Analysis}
Quantitative detection results are presented in Tab.~\ref{tab:Quantitative_Detection}, which lists the mean average precision (mAP) under different intersection over union (IoU) thresholds. The results indicate that our method achieves superior performance in mAP@[0.5:0.95] compared to both source images and other comparative methods, demonstrating consistently strong mAP performance across varying IoU criteria. These findings confirm that our approach not only significantly boosts downstream task performance relative to single-modality detection but also substantially outperforms state-of-the-art fusion algorithms in real-world detection scenarios.

\section{Conclusion} \label{sec:conclusion}
In this paper, we propose WIFE-Fusion, a novel multi-modal image fusion framework that employs effective cross-modality and cross-frequency interactions to fully leverage complementary characteristics and extract structural and detail information from source images. The Intra-Frequency Self-Attention (IFSA) module utilizes shared self-attention within each frequency band to explore inter-modal relevant and complementary information, while the Inter-Frequency Interaction (IFI) module enables cross-modal exchanges between different frequency components, enhancing salient features and excavating latent information. Comprehensive experiments across three multimodal fusion tasks demonstrate that WIFE-Fusion surpasses both task-specific and general-purpose fusion methods in achieving superior image fusion quality.

\bibliographystyle{IEEEtran}

\bibliography{reference}

\end{document}